%% file: main.tex
\documentclass[final]{cvpr}

\usepackage{times}
\usepackage{epsfig}
\usepackage{graphicx}
\usepackage{amsmath}
\usepackage{amssymb}

\usepackage{comment}
\usepackage{enumitem}
\usepackage{color}
\usepackage{makecell}
\usepackage{multirow}
\usepackage[nocompress,sort]{cite}
\usepackage{widetable}

\newcommand\customparagraph[1]{\vspace{0.4em}\noindent\textbf{#1.}}

\newcommand{\Sec}{\textit{Sec}.\@\xspace}

\input{tex/hyperref}

\usepackage[labelfont=bf,
            format=plain]{caption}

\begin{document}

\input{tex/title}

\input{tex/abstract}
\input{tex/introduction}
\input{tex/related_work}

\input{tex/method}
\input{tex/experiments}
\input{tex/conclusion}

\input{tex/appendix}

{\small
\bibliographystyle{ieee_fullname}
\bibliography{main}
}

\end{document}

%% file: tex/hyperref.tex
\usepackage[pagebackref=true]{hyperref}

\definecolor{linkcolour}{rgb}{0.949,0.419,0.541}    %
\definecolor{citecolour}{rgb}{0.000,0.600,0.200}  %
\definecolor{urlcolour} {rgb}{0.078,0.337,0.564}    %

\hypersetup{
    bookmarksopen=false,
    bookmarksnumbered=true,
    pdfpagemode=UseOutlines,  %
    colorlinks=true,          %
    linkcolor=linkcolour,     %
    anchorcolor=black,        %
    citecolor=citecolour,     %
    filecolor=urlcolour,      %
    menucolor=blue,           %
    urlcolor=urlcolour,       %
    breaklinks=true,          %
    pdfstartview={FitBH},
    pdfview={FitBH},
}

%% file: tex/title.tex
\title{PatchmatchNet: Learned Multi-View Patchmatch Stereo}

\author{Fangjinhua Wang$^1$
        \quad
        Silvano Galliani$^2$
        \quad
        Christoph Vogel$^2$
        \quad
        Pablo Speciale$^2$
        \quad
        Marc Pollefeys$^{1,2}$\\
        $^1$Department of Computer Science, ETH Zurich\\
        $^2$Microsoft Mixed Reality \& AI Zurich Lab}
\maketitle

%% file: tex/abstract.tex
\begin{abstract}
We present PatchmatchNet, a novel and learnable cascade formulation of Patchmatch for high-resolution multi-view stereo.
With high computation speed and low memory requirement, PatchmatchNet can process higher resolution imagery and is more suited to run on resource limited devices than competitors that employ 3D cost volume regularization.
For the first time we introduce an iterative multi-scale Patchmatch in an end-to-end trainable 
architecture and improve the Patchmatch core algorithm with a 
novel and learned adaptive propagation and evaluation scheme for each iteration. 
Extensive experiments show a very competitive performance and generalization for our method on 
DTU, Tanks \& Temples and ETH3D, 
but at a significantly higher efficiency than all existing top-performing models:
at least two and a half times faster than state-of-the-art methods with twice less memory usage. Code is available at \url{https://github.com/FangjinhuaWang/PatchmatchNet}.

\end{abstract}

%% file: tex/introduction.tex
\section{Introduction}
Given a collection of images with known camera parameters, multi-view stereo (MVS) 
describes the task of reconstructing the dense geometry of the observed scene. 
Despite being a fundamental problem of geometric computer vision that has been studied for several decades, 
MVS is still a challenge. This is due to a variety of de-facto unsolved problems occurring in practice 
such as occlusion, illumination changes, untextured areas and non-Lambertian 
surfaces~\cite{aanaes2016_dtu,knapitsch2017tanks,2017eth3d}. %

\begin{figure}
\vspace{-0.2cm}
\setlength{\belowcaptionskip}{-0.7cm}
\centering
\setlength{\abovecaptionskip}{0.2cm}
{\includegraphics[width=1.0\linewidth]{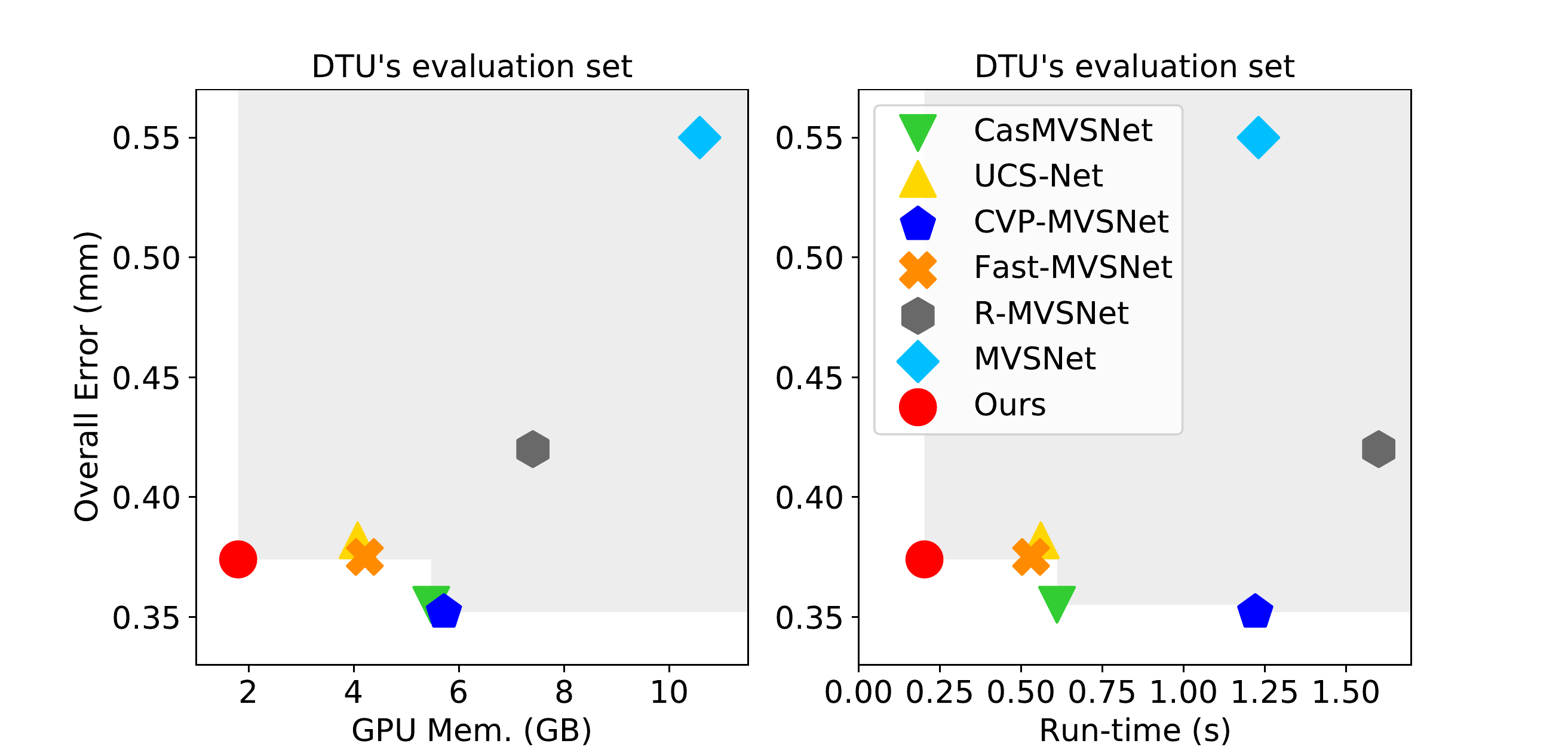}}
\caption{
Comparison with state-of-the-art learning-based multi-view stereo 
methods~\cite{yao_2018_mvsnet,yao_2019_rmvsnet,yu_2020_fastmvsnet,gu_2020_cascademvsnet,cheng_2020_ucsnet,yang_2020_cvpr} 
on DTU~\cite{aanaes2016_dtu}. 
Relationship between error, GPU memory  
and run-time with image size $1152\!\times\!864$.
}\label{fig:teaser}
\end{figure}

The success of Convolutional Neural Networks (CNN) in almost any field of computer vision ignites 
the hope that data driven models can solve some of these issues that classical MVS models struggle with. 
Indeed, many learning-based methods~\cite{yao_2018_mvsnet, yao_2019_rmvsnet, xu2020learning_inverse,luo_2019_p_mvsnet,chen_2019_pointmvsnet} appear to fulfill such promise and outperform some traditional methods~\cite{schoenberger2016colmap, galliani_2015_gipuma} 
on MVS benchmarks~\cite{aanaes2016_dtu,knapitsch2017tanks}. 
While being successful at the benchmark level, most of them do 
only pay limited attention to scalability, memory and run-time. 
Currently, most learning-based MVS methods~\cite{yao_2018_mvsnet,xu2020learning_inverse,luo_2019_p_mvsnet,chen_2019_pointmvsnet} 
construct a 3D cost volume, regularize it with a 3D CNN and regress the depth. As 3D CNNs are usually time and memory consuming, some methods~\cite{yao_2018_mvsnet,xu2020learning_inverse} down-sample 
the input during feature extraction and compute both, the cost volume and the depth map at low-resolution. 
Yet, according to Fig.~\ref{fig:teaser}, delivering depth maps at low resolution can harm accuracy. 
Methods that do not scale up well to realistic image sizes of several mega-pixel 
cannot exploit the full resolution due to memory limitations. 
Evidently, low memory and time consumption are key to enable processing on 
memory and computational restricted devices such as phones or mixed reality headsets, as well as in time critical applications. 
Recently, researchers tried to alleviate these 
limitations. For example, R-MVSNet~\cite{yao_2019_rmvsnet} decouples the memory requirements from the depth range %
and sequentially processes the cost volume at the cost of an additional runtime penalty. ~\cite{gu_2020_cascademvsnet,cheng_2020_ucsnet,yang_2020_cvpr} include cascade 3D cost volumes to predict high-resolution depth map from coarse to fine with high efficiency in time and memory.

Several traditional MVS methods~\cite{galliani_2015_gipuma, zheng_2014_cvpr, schoenberger2016colmap, xu_2019_acmm} 
abandon the idea of holding a structured cost volume completely and instead 
are based on the seminal Patchmatch~\cite{barnes_2009_patchmatch} algorithm. 
Patchmatch adopts a randomized, iterative algorithm for approximate nearest neighbor field computation~\cite{barnes_2009_patchmatch}. 
In particular, the inherent spatial coherence of depth maps is exploited to quickly find a good solution 
without the need to look through all possibilities. 
Low memory requirements -- independent of the disparity range -- and an implicit smoothing effect 
make this method very attractive for our deep learning based MVS setup. 

In this work, we propose PatchmatchNet, a novel cascade formulation of learning-based Patchmatch, 
which aims at decreasing memory consumption and run-time for high-resolution multi-view stereo. 
It inherits the advantages in efficiency from classical Patchmatch, 
but also aims to improve the performance with the power of deep learning.

\vspace{0.4em}\noindent\textbf{Contributions:}
\textbf{(\romannumeral 1)}
We introduce the Patchmatch idea into an end-to-end trainable deep learning based MVS framework. 
Going one step further, we embed the model into a coarse-to-fine framework to speed up computation.
\textbf{(\romannumeral 2)}
We augment the traditional propagation and cost evaluation steps of Patchmatch with learnable, 
adaptive modules that improve accuracy and base both steps on deep features. 
We estimate visibility information during cost aggregation for the source views. 
Moreover, we propose a robust training strategy to introduce randomness into training for improved robustness in visibility estimation and generalization. 
\textbf{(\romannumeral 3)}
We verify the effectiveness of our method on various MVS  datasets,~\eg 
DTU~\cite{aanaes2016_dtu}, Tanks \& Temples~\cite{knapitsch2017tanks} and ETH3D~\cite{2017eth3d}. 
The results demonstrate that our PatchmatchNet achieves competitive performance, 
while reducing memory consumption and run-time compared to most learning-based methods.

%% file: tex/related_work.tex
\section{Related Work}
\customparagraph{Traditional MVS}
Traditional MVS methods can be divided into four categories: 
voxel-based~\cite{sinha2007, ulusoy2017}, surface evolution based~\cite{furukawa2006, li2016}, 
patch-based~\cite{furukawa2010, locher_2016_cvpr} and depth map based~\cite{schoenberger2016colmap, galliani_2015_gipuma,xu_2019_acmm}. 
Comparatively, depth map based methods are more concise and flexible. 
Here, we discuss Patchmatch Stereo methods~\cite{galliani_2015_gipuma, schoenberger2016colmap, xu_2019_acmm} in this category. %
Galliani~\etal~\cite{galliani_2015_gipuma} present Gipuma, a massively parallel multi-view extension of Patchmatch stereo. 
It uses a red-black checkerboard pattern to parallelize message-passing during propagation.  %
Sch\"{o}nberger~\etal~\cite{schoenberger2016colmap} present COLMAP, which jointly estimates pixel-wise view selection, depth map and surface normal. 
ACMM~\cite{xu_2019_acmm} adopts adaptive checkerboard sampling, multi-hypothesis joint view selection and multi-scale geometric consistency guidance. 
Based on the idea of Patchmatch, we propose our learning-based Patchmatch, 
which inherits the efficiency from classical Patchmatch, but also improves the performance leveraging deep learning. 

\customparagraph{Learning-based stereo}
GCNet~\cite{kendall_2017_gcnet} introduces 3D cost volume regularization for stereo estimation and regresses the final disparity map with a soft argmin operation. 
PSMNet~\cite{chang_2018_psmnet} adds spatial pyramid pooling (SPP) and applies a 3D hour-glass network for regularization. 
DeepPruner~\cite{duggal2019deeppruner} develops a differentiable Patchmatch module, without learnable parameters, to discard most disparities and then builds a lightweight cost volume, 
which is regularized by a 3D CNN. 
In contrast, we do not apply any cost volume regularization but extend the original Patchmatch idea into the deep learning era. 
Xu~\etal~\cite{xu_2020_aanet} propose a sparse point based intra-scale cost aggregation
method with deformable convolution~\cite{dai_2017_deformconv}. %
Likewise, we propose a strategy to adaptively sample points for spatial cost aggregation. 

\customparagraph{Learning-based MVS}
Voxel-based methods~\cite{ji_2017_surfacenet,kar2017learning} are restricted to small-scale reconstructions, 
due to the drawbacks of a volumetric representation. 
In contrast, based on plane-sweep stereo~\cite{planesweeping}, many recent works~\cite{yao_2018_mvsnet,chen_2019_pointmvsnet,luo_2019_p_mvsnet,xu2020learning_inverse} use depth maps to reconstruct the scene. %
They build cost volumes with warped features from multiple views, regularize them with 3D CNNs and regress the depth. As 3D CNNs are time and memory consuming, they usually use down-sampled cost volumes. 
To reduce memory, R-MVSNet~\cite{yao_2019_rmvsnet} sequentially regularizes 2D cost maps with GRU~\cite{cho2014learning} but sacrifices run-time.
Current research targets to improve efficiency and also estimate high-resolution depth maps. 
CasMVSNet~\cite{gu_2020_cascademvsnet} proposes cascade cost volumes based on a feature pyramid and estimates the depth map in a coarse-to-fine manner. 
UCS-Net~\cite{cheng_2020_ucsnet} proposes cascade adaptive thin volumes, which use variance-based uncertainty estimates for an adaptive construction. 
CVP-MVSNet~\cite{yang_2020_cvpr} forms an image pyramid and also constructs a cost volume pyramid. 
To accelerate propagation in Patchmatch, we likewise employ a cascade formulation. 
In addition to cascade cost volumes, PVSNet~\cite{xu2020pvsnetpv} learns to predict visibility for each source image. 
An anti-noise training strategy is used to introduce disturbing views. 
We also learn a strategy to adaptively combine the information of multiple views based on visibility information. Moreover, we propose a robust training strategy to include randomness into the training to improve robustness in visibility estimation and generalization. 
Fast-MVSNet~\cite{yu_2020_fastmvsnet} constructs a sparse cost volume to learn a sparse depth map and then use high-resolution RGB image and  2D CNN to densify it. 
We build a refinement module and use the RGB image to guide the up-sampling of the depth map based on MSG-Net~\cite{hui16msgnet}.

%% file: tex/method.tex
\def\negativeCaptionSpace{\vspace{-0.1cm}} %
\begin{figure}[htbp]
\vspace{-0.2cm}
\setlength{\belowcaptionskip}{-0.3cm}
\centering
\setlength{\abovecaptionskip}{0.2cm}
{\includegraphics[width=\linewidth]{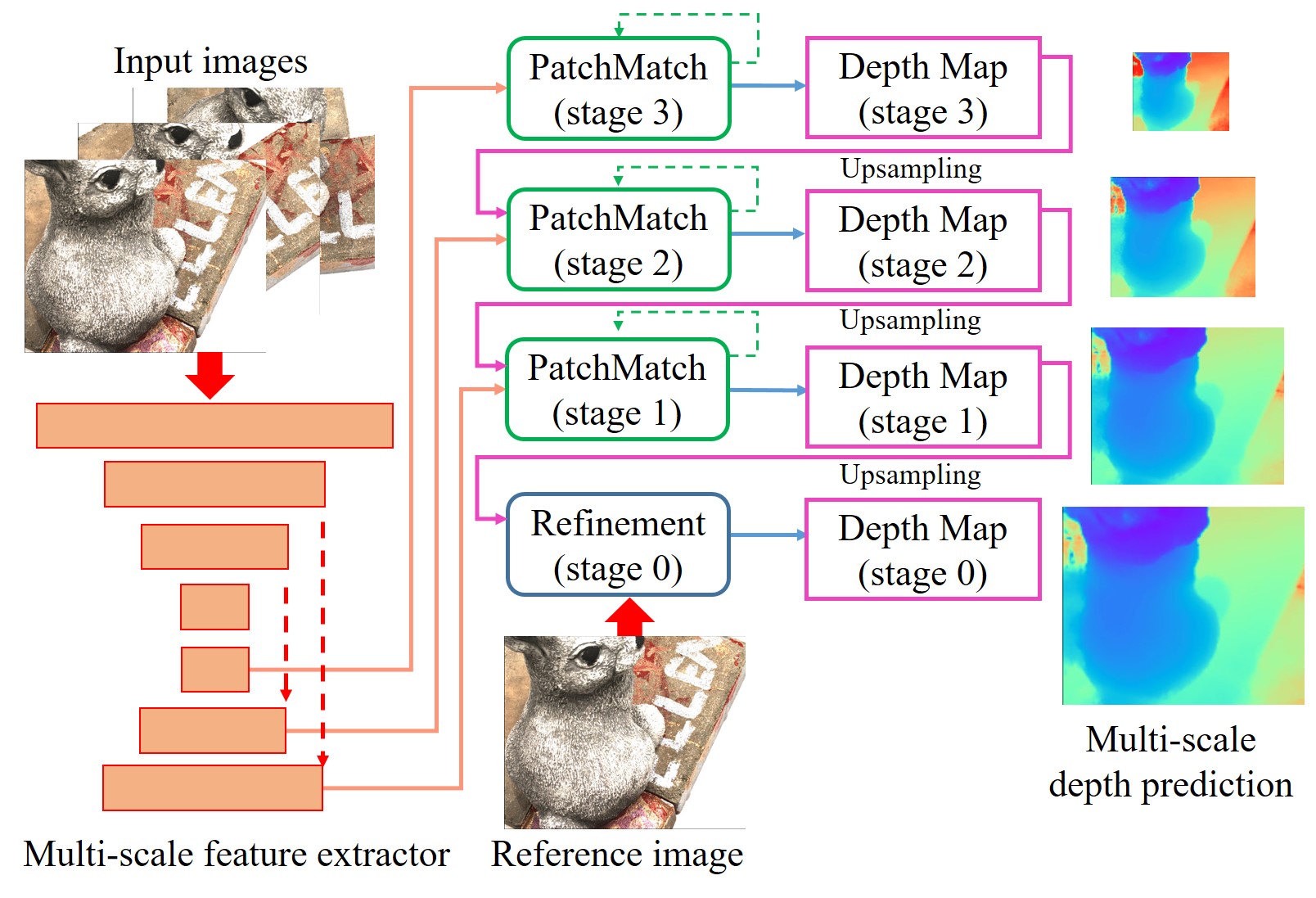}}
\caption{Structure of PatchmatchNet: multi-scale feature extractor, learning-based Patchmatch and refinement. Patchmatch is applied for multiple iterations on several stages to predict the depth map in a coarse-to-fine manner. %
Refinement uses the input to guide upsampling of the final depth map. 
On stage $k$, the resolution of the depth maps is $\frac{W}{2^k}\!\times\!\frac{H}{2^k}$, with input images of size $W\!\times\!H$.
}
\label{fig:overall_structure}
\end{figure}
\section{Method}
In this section, we introduce the structure of PatchmatchNet, illustrated in Fig.~\ref{fig:overall_structure}.
It consists of multi-scale feature extraction, learning-based Patchmatch included iteratively in a coarse-to-fine framework, and a spatial refinement module.

\subsection{Multi-scale Feature Extraction}
Given $N$ input images of size $W \!\times\! H$, we use $\mathbf{I}_0$ and ${\left\{\mathbf{I}_i\right\}}_{i=1}^{N-1}$ to denote reference and source images respectively. 
Before we apply our learning-based Patchmatch algorithm, we extract pixel-wise features from our inputs, 
similar to Feature Pyramid Network (FPN)~\cite{lin2017fpn}. 
Features are extracted hierarchically at multiple resolutions, which allows us to advance our depth map estimation in a coarse-to-fine manner.
 
\subsection{Learning-based Patchmatch}
Following traditional Patchmatch~\cite{barnes_2009_patchmatch} and subsequent adaptations to depth map 
estimation~\cite{bleyer2011patchmatch,galliani_2015_gipuma}, our learnable Patchmatch consists of the following three main steps:

\begin{itemize}[topsep=1ex]
  \setlength\itemsep{0.1em}
  \item [1.]Initialization: generate random hypotheses;
  \item [2.]Propagation: propagate hypotheses to neighbors;
  \item [3.]Evaluation: compute the matching costs for all the hypotheses and choose best solutions.
\end{itemize}

\begin{figure*}[htbp]

\setlength{\belowcaptionskip}{-0.5cm}
\centering
\setlength{\abovecaptionskip}{0.2cm}
{\includegraphics[width=0.9\linewidth]{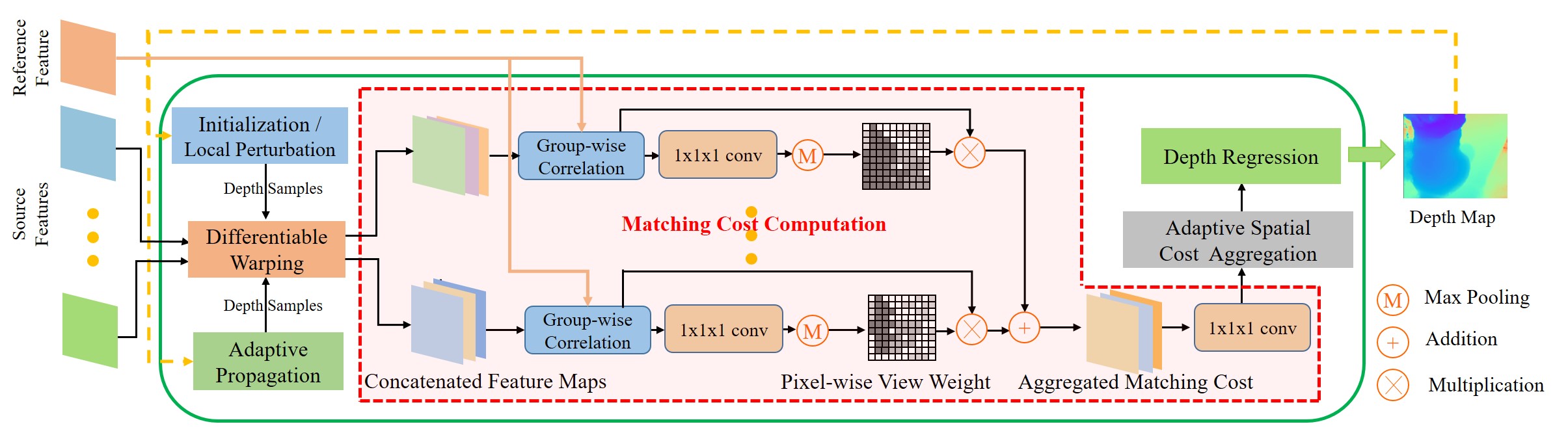}}
\caption{
Detailed structure of learned Patchmatch. 
At the initial iteration of coarsest stage 3 only random depth hypotheses in \emph{initialization} are used.
Afterwards, hypotheses are obtained from \emph{adaptive propagation} and 
\emph{local perturbation}, the latter providing depth samples around the previous estimate. 
The learned pixel-wise view weight is estimated in the first iteration of Patchmatch and kept fixed in the matching cost computation.
}
\label{fig:patchmatch}
\end{figure*}

After initialization, the approach iterates between propagation and evaluation until convergence.
Leveraging deep learning, we propose an adaptive version of the propagation (\Sec\ref{adaptive_propagation})
and evaluation (\Sec\ref{evaluation}) module and also adjust the initialization (\Sec\ref{initialization}). 
The detailed structure of our Patchmatch pipeline is illustrated in Fig.~\ref{fig:patchmatch}. 
In a nutshell, the propagation module adaptively samples the points for propagation based on the extracted deep features.
Our adaptive evaluation learns to estimate visibility information for cost computation and
adaptively samples the spatial neighbors to aggregate the costs again based on deep features. 
Unlike \cite{bleyer2011patchmatch, galliani_2015_gipuma, xu_2019_acmm}, we refrain from
parameterizing the per-pixel hypothesis as a slanted plane, due to heavy memory penalties. %
Instead, we rely on our learned adaptive evaluation to organize the 
spatial pattern within the window over which matching costs are computed.

\subsubsection{Initialization and Local  Perturbation}\label{initialization}
In the very first iteration of Patchmatch, the initialization is performed in a random manner to promote diversification. 
Based on a pre-defined depth range $[d_{min}, d_{max}]$, 
we sample per pixel $D_f$ depth hypotheses in the \emph{inverse} depth range, corresponding to uniform sampling in image space. This helps our model be applicable to complex and large-scale scenes~\cite{xu2020learning_inverse,yao_2019_rmvsnet}. To ensure we cover the depth range evenly, we divide the (inverse) range into $D_f$ intervals and ensure that each interval is covered by one hypothesis.

For subsequent iterations on stage $k$, we perform local perturbation by generating per pixel $N_k$ hypotheses uniformly
in the normalized inverse depth range $R_k$
and gradually decrease $R_k$ for finer stages. %
To define the center of $R_k$, we utilize the estimation from previous iteration, 
possibly up-sampled from a coarser stage.
This delivers a more diverse
set of hypotheses than just using propagation. %
Sampling around the previous estimation can refine the result locally and correct wrong estimates (see supplementary).

\vspace{-0.2cm}
\subsubsection{Adaptive Propagation} \label{adaptive_propagation}

\begin{figure}
\centering
\setlength{\belowcaptionskip}{-0.5cm}
\setlength{\abovecaptionskip}{0.1cm}
{\includegraphics[width=1\linewidth]{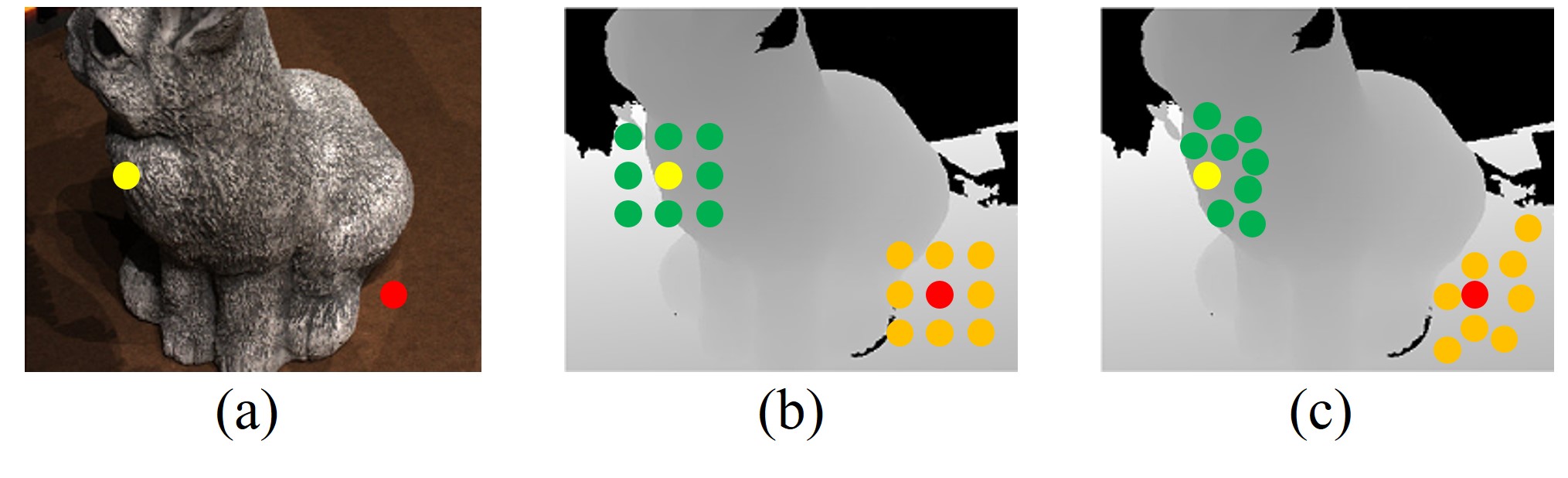}}
\caption{
Sampled locations with adaptive propagation.
Pixels located at the object boundary (yellow) and a textureless region (red) receive depth hypotheses from sampled neighbors (green and orange).
(a) Reference image. %
(b) Fixed sampling locations of classic propagation. 
(c) Adaptive sampling locations with adaptive propagation. The grayscale image in (b) and (c) is the ground truth depth map.
}
\label{fig:propagation}
\end{figure}

Spatial coherence of depth values does in general only exist for pixel from the same physical surface.
Hence, instead of propagating depth hypotheses naively from a static set of neighbors %
as done for Gipuma~\cite{galliani_2015_gipuma} and DeepPruner~\cite{duggal2019deeppruner},
we want to perform the propagation in an adaptive manner, which gathers hypotheses from the same surface.
This helps Patchmatch converge faster and deliver more accurate depth maps. 
Fig.~\ref{fig:propagation} illustrates the idea and functionality of our strategy. 
Our adaptive scheme tends to gather hypotheses from pixels of the same surface -- 
for both the textured object and the textureless region -- 
enabling us to effectively collect more promising depth hypotheses 
compared to using just a static pattern.

We base our implementation of the adaptive propagation on Deformable Convolution Networks~\cite{dai_2017_deformconv}. 
As the approach is identical for each resolution stage, we omit subindices denoting the stage. %
To gather $K_{p}$ depth hypotheses for pixel $\mathbf{p}$ in the reference image, 
our model learns additional 2D offsets ${\left\{\Delta \mathbf{o}_i(\mathbf{p})\right\}}_{i=1}^{K_{p}}$ that are applied on top of 
fixed 2D offsets ${\left\{\mathbf{o}_i\right\}}_{i=1}^{K_{p}}$, 
organized as a grid. %
We apply a 2D CNN on the reference feature map $\mathbf{F}_0$ to learn additional 2D offsets for each pixel $\mathbf{p}$ and get the depth hypotheses $\mathbf{D}_{p}(\mathbf{p})$  via bilinear interpolation as follows: 
\begin{equation}
\setlength{\abovedisplayskip}{3pt}
\setlength{\belowdisplayskip}{3pt}
    \mathbf{D}_{p}(\mathbf{p}) 
    = \left\{\mathbf{D}(\mathbf{p} + \mathbf{o}_i + \Delta \mathbf{o}_i(\mathbf{p})) \right\}_{i=1}^{K_{p}},
\end{equation}
where $\mathbf{D}$ is the depth map from previous iteration, possibly up-sampled from a coarser stage. %

\subsubsection{Adaptive Evaluation} \label{evaluation}
The adaptive evaluation module performs the following steps:
differentiable warping, matching cost computation, adaptive spatial cost aggregation and depth regression. 
As the approach is identical at each resolution stage, we omit subindices to ease notation.

\customparagraph{Differentiable Warping}
Following plane sweep stereo~\cite{planesweeping},
most learning-based MVS methods~\cite{cheng_2020_ucsnet,gu_2020_cascademvsnet,luo_2019_p_mvsnet, yao_2018_mvsnet, yao_2019_rmvsnet} 
establish front-to-parallel planes at sampled depth hypotheses and warp the feature maps of source images into them. 
Equipped with intrinsic matrices $\left\{K_i\right\}_{i=0}^{K}$ and relative 
transformations $\left\{\left[\mathbf{R}_{0,i}|\mathbf{t}_{0,i} \right] \right\}_{i=1}^{K}$ 
of reference view $0$ and source view $i$, 
we compute the corresponding pixel $\mathbf{p}_{i,j}:=\mathbf{p}_{i}(d_j)$ in the source
for a pixel $\mathbf{p}$ in the reference, given in homogeneous coordinates, 
and depth hypothesis $d_j:=d_j(\mathbf{p})$ as follows:
\begin{equation}
\setlength{\abovedisplayskip}{3pt}
\setlength{\belowdisplayskip}{6pt}
    \mathbf{p}_{i,j} = \mathbf{K}_i \cdot (\mathbf{R}_{0,i}\cdot(\mathbf{K}_0^{-1} \cdot \mathbf{p} \cdot d_j) + \mathbf{t}_{0,i}).
\end{equation}
We obtain the warped source feature maps of view $i$ and the $j$-th set 
of (per pixel different) depth hypotheses, 
$\mathbf{F}_i(\mathbf{p}_{i,j})$, %
via differentiable bilinear interpolation. 

\customparagraph{Matching Cost Computation}
For multi-view stereo, this step has to integrate information from %
an arbitrary number of %
source views into a single cost per pixel $\mathbf{p}$ and depth hypothesis $d_j$. 
To that end, we compute the cost per hypothesis via group-wise correlation~\cite{xu2020learning_inverse} 
and aggregate over the views with a pixel-wise view weight~\cite{schoenberger2016colmap,xu_2019_acmm,xu2020pvsnetpv}.
In that manner we can employ visibility information during cost aggregation and gain robustness. 
Finally, the per group costs are projected into a single number, per reference pixel and hypothesis, by a small network. 

Let $\mathbf{F}_0(\mathbf{p}),\,\mathbf{F}_i(\mathbf{p}_{i,j})\!\in\!\mathbb{R}^C$
be the features in the reference and source feature maps respectively.
After dividing their feature channels evenly into $G$ groups, 
$\mathbf{F}_0(\mathbf{p})^g$ and $\mathbf{F}_i(\mathbf{p}_{i,j})^g$, 
the $g$-th group similarity %
$\mathbf{S}_{i}(\mathbf{p},j)^g\in\mathbb{R}$ is computed as:
\begin{equation}
\setlength{\abovedisplayskip}{3pt}
\setlength{\belowdisplayskip}{3pt}
    \mathbf{S}_{i}(\mathbf{p},j)^g = \frac{G}{C} \left \langle \mathbf{F}_0(\mathbf{p})^g, \mathbf{F}_i(\mathbf{p}_{i,j})^g\right \rangle,
\end{equation}
where $\left \langle \cdot, \cdot \right \rangle$ is the inner product.
We use $\mathbf{S}_{i}(\mathbf{p},j)\in\mathbb{R}^G$ to denote the respective group similarity vector. 
Agglomeration over hypotheses and pixels delivers the tensor $\mathbf{S}_i\in\mathbb{R}^{W \times H \times D \times G}$.

To find pixel-wise view weights, ${\left\{\mathbf{w}_i(\mathbf{p})\right\}}_{i=1}^{N-1}$,
we exploit the diversity of our initial set of depth hypotheses in the first iteration on stage 3 (\Sec\ref{initialization}).
We intend $\mathbf{w}_i(\mathbf{p})$ to represent the visibility information of pixel $\mathbf{p}$ in the source image $\mathbf{I}_i$. 
The weights are computed once and kept fixed and up-sampled for finer stages. 

A simple pixel-wise view weight network, 
composed of 3D convolution layers with $1 \!\times\! 1 \!\times\! 1$ kernels and sigmoid non-linearities, 
takes the initial set of similarities $\mathbf{S}_i$ to output a number between $0$ and $1$ per pixel and depth hypothesis to produce $\mathbf{P}_{i}\!\in\!\mathbb{R}^{W\times H\times D}$. 
The view weights for pixel $\mathbf{p}$ and source image $\mathbf{I}_i$ are given by:
\begin{equation}
\setlength{\abovedisplayskip}{3pt}
\setlength{\belowdisplayskip}{3pt}
    \mathbf{w}_i(\mathbf{p}) = \max \left\{ \mathbf{P}_{i}(\mathbf{p},j) | j=0,1,\dots,D-1 \right\},
\end{equation}
where $\mathbf{P}_{i}(\mathbf{p},j)$ intuitively represents confidence of visibility  for the range covered by the $j$-th depth hypothesis at $\mathbf{p}$. 

The final per group similarities $\mathbf{\bar{S}}(\mathbf{p},j)$ for pixel $\mathbf{p}$ and the $j$-th hypothesis
are the weighted sum of $\mathbf{S}_{i}(\mathbf{p},j)$ and the view weight $\mathbf{w}_i(\mathbf{p})$:
\begin{equation}
\setlength{\abovedisplayskip}{3pt}
\setlength{\belowdisplayskip}{3pt}
    \mathbf{\bar{S}}(\mathbf{p},j) = \frac{\sum_{i=1}^{N-1}\mathbf{w}_i(\mathbf{p}) \cdot \mathbf{S}_{i}(\mathbf{p},j)}{\sum_{i=1}^{N-1}\mathbf{w}_i(\mathbf{p})}.
\end{equation}

Finally, we compose $\mathbf{\bar{S}}(\mathbf{p},j)$ for all pixels and hypotheses into 
$\mathbf{\bar{S}}\in\mathbb{R}^{W \times H \times D \times G}$ 
and apply a small network with 3D convolution and $1 \!\times\! 1 \!\times\! 1$ kernels 
to obtain a single cost, 
$\mathbf{C}\in\mathbb{R}^{W \times H \times D}$, %
per pixel and depth hypothesis.

\begin{figure}
\vspace{-0.2cm}
\setlength{\belowcaptionskip}{-0.7cm}
\centering
\setlength{\abovecaptionskip}{0.2cm}
{\includegraphics[width=1\linewidth]{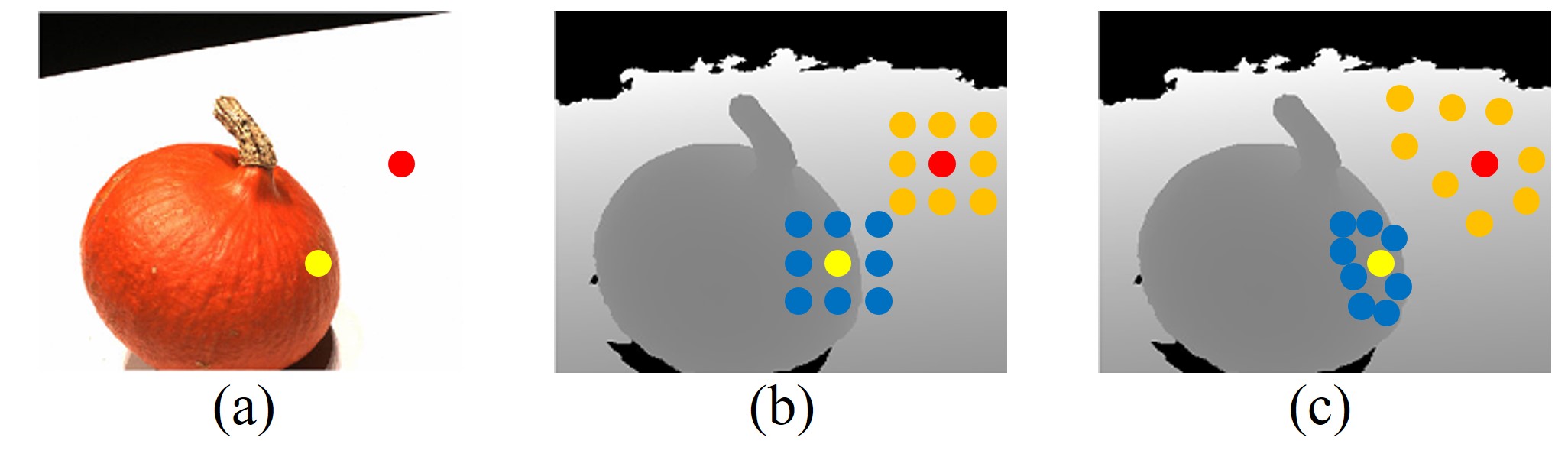}}
\caption{
    Sampled locations in adaptive spatial cost aggregation.
    Pixels located at an object boundary (yellow) and a textureless area (red) 
    aggregate matching costs from sampled neighbors (blue and orange). 
    (a) Reference image for depth prediction. 
    (b) Fixed sampling locations. 
    (c) Adaptive sampling locations of our method. 
    The grayscale image in (b) and (c) is the ground truth depth map.
    }
\label{fig:evaluation}
\end{figure}

\customparagraph{Adaptive Spatial Cost Aggregation}
Traditional MVS matching algorithms often aggregate costs over a spatial window (\ie in our case a front-to-parallel plane) 
for increased matching robustness and an implicit smoothing effect. 
Arguably, our multi-scale feature extractor already aggregates neighboring information from a large receptive field in the spatial domain. 
Nevertheless, we propose to look into spatial cost aggregation. 
To prevent the problem of aggregating across surface boundaries, we propose an adaptive spatial aggregation strategy
based on Patchmatch stereo~\cite{bleyer2011patchmatch} and AANet~\cite{xu_2020_aanet}. 
For a spatial window of $K_{e}$ pixels $\left\{ \mathbf{p}_k \right\}_{k=1}^{K_{e}}$ are organized as a grid,
we learn per pixel additional offsets $\left\{ \Delta \mathbf{p}_k \right\}_{k=1}^{K_{e}}$. 
The aggregated spatial cost $\tilde{\mathbf{C}}(\mathbf{p},j)$ is defined as:
\begin{equation}\label{cost aggregation}
\setlength{\abovedisplayskip}{3pt}
\setlength{\belowdisplayskip}{3pt}
    \tilde{\mathbf{C}}(\mathbf{p},j) = \frac{1}{\sum_{k=1}^{K_{e}} \!w_k d_k}\! \sum_{k=1}^{K_{e}} w_k d_k  \mathbf{C}(\mathbf{p}\!+\!\mathbf{p}_k\!+\!\Delta \mathbf{p}_k,j),
\end{equation}
where $w_k$ and $d_k$ weight the cost $\mathbf{C}$ based on feature and depth similarity (details in supplementary). 
Similar to adaptive propagation, 
the per pixel sets of displacements $\left\{ \Delta \mathbf{p}_k \right\}_{k=1}^{K_{e}}$ are
found by applying a 2D CNN on the reference feature map $\mathbf{F}_0$. 
Fig.~\ref{fig:evaluation} exemplifies the learned adaptive aggregation window. 
Sampled locations stay within object boundaries, while for the textureless region, 
the sampling points aggregate over a larger spatial context, 
which can potentially reduce the ambiguity of estimation.

\customparagraph{Depth Regression}
Using \emph{softmax} we turn the (negative) cost $\tilde{\mathbf{C}}$ to a probability $\mathbf{P}$,
which is used for sub-pixel depth regression and measure estimation confidence~\cite{yao_2018_mvsnet}. 
The regressed depth value $\mathbf{D}(\mathbf{p})$ at pixel $\mathbf{p}$ is found 
as the expectation \wrt $\mathbf{P}$ of the hypotheses: 
\begin{equation}
\setlength{\abovedisplayskip}{3pt}
\setlength{\belowdisplayskip}{3pt}
    \mathbf{D}(\mathbf{p}) = \sum_{j=0}^{D-1} d_j \cdot \mathbf{P}(\mathbf{p},j).
\end{equation}

\subsection{Depth Map Refinement}
Instead of using Patchmatch also on the finest resolution level (stage 0), 
we find it sufficient to directly up-sample (from resolution $\frac{W}{2} \!\times\! \frac{H}{2}$ to $W\!\times\! H$) 
and refine our estimation with the RGB image.
Based on MSG-Net~\cite{hui16msgnet}, we design a depth residual network. 
To avoid being biased for a certain depth scale, we pre-scale the input depth map into the range $[0,1]$ and convert it back after refinement. 
Our refinement network learns to output a residual that is added to the (up-sampled) estimation from 
Patchmatch, $\mathbf{D}$, to get the refined depth map $\mathbf{D}_{ref}$. 
This network independently extracts feature maps $\mathbf{F}_D$ and $\mathbf{F}_I$ from $\mathbf{D}$ and $\mathbf{I}_0$ 
and applies deconvolution on $\mathbf{F}_D$ to up-sample the feature map to the image size. 
Multiple 2D convolution layers are applied on top of the concatenation of both feature maps -- depth map and image -- to deliver the depth residual.

\subsection{Loss Function}
Loss function $L_{total}$ considers the losses among all the %
depth estimation and rendered ground truth with same resolution as a sum: 
\begin{equation}
\setlength{\abovedisplayskip}{3pt}
\setlength{\belowdisplayskip}{3pt}
    L_{total} = \sum_{k=1}^3 \sum_{i=1}^{n_k} L_i^k + L_{ref}^0.
\end{equation}
We adopt the smooth $L1$ loss for $L_i^k$, the loss of the $i$-th iteration of Patchmatch on stage $k$ ($k=1,2,3$) 
and $L_{ref}^0$, the loss for final refined depth map. 

%% file: tex/experiments.tex
\def\negativeCaptionSpace{\vspace{-0.0cm}} %
\section{Experiments}
We evaluate our work on multiple datasets, such as DTU~\cite{aanaes2016_dtu}, Tanks \& Temples~\cite{knapitsch2017tanks} and ETH3D~\cite{2017eth3d} and analyze each new component with an ablation study.

\subsection{Datasets}
The DTU dataset~\cite{aanaes2016_dtu} is an indoor multi-view stereo dataset with 124 different scenes where all scenes share the same camera trajectory. %
We use the training, testing and validation split introduced in~\cite{ji_2017_surfacenet}.
The Tanks \& Temples dataset~\cite{knapitsch2017tanks} is provided as a set of video sequences in realistic environments. 
It is divided into intermediate and advanced datasets. %
ETH3D benchmark~\cite{2017eth3d} consists of calibrated high-resolution images of scenes with strong viewpoint variations. 
It is divided into training and test datasets. %

\subsection{Robust Training Strategy} \label{robust training}
Many learning-based methods~\cite{yao_2018_mvsnet, luo_2019_p_mvsnet, xu2020learning_inverse, gu_2020_cascademvsnet, cheng_2020_ucsnet, yang_2020_cvpr} select two best source views based on view selection scores~\cite{yao_2018_mvsnet} 
to train models on DTU~\cite{aanaes2016_dtu}. 
However, the selected source views have a strong visibility correlation with the reference view, 
which may affect the training of the pixel-wise view weight network.
Instead, we propose a robust training strategy based on PVSNet~\cite{xu2020pvsnetpv}.
For each reference view, we randomly choose four from the ten best source views for training. 
This strategy increases the diversity at training time and augments the dataset on the fly, which improves the generalization performance.
In addition, training on those random source views with weak visibility correlation generates further robustness for our visibility estimation. 

\subsection{Implementation Details} \label{implementation}
We implement the model with PyTorch~\cite{paszke2019pytorchai} and train it on DTU's training set~\cite{aanaes2016_dtu}. 
We set the image resolution to $640 \times 512$ and the number of input images to $N=5$.
The selection of source images is based on the proposed robust training strategy.
We set the iteration number of Patchmatch on stages $3,2,1$ as $2,2,1$. 
For initialization, we set $D_f\!=\!48$. For local  perturbation, we set  $R_3\!=\!0.38,R_2\!=\!0.09,R_1\!=\!0.04$ (see supplementary),  $N_3\!=\!16,N_2\!=\!8,N_1\!=\!8$. 
In the adaptive propagation, we set $K_{p}$ on stages $3,2,1$ to 16, 8, 0 (no propagation for last iteration on stage 1, see supplementary). 
For the adaptive evaluation, we use $K_{e}=9$ on all stages. %
We train our model with Adam~\cite{kingma2015adamam} ($\beta_1 \!=\! 0.9, \beta_2 \!=\! 0.999$) for 8 epochs with a learning rate of 0.001. 
Here, we use a batch size of 4 and train on 2 Nvidia GTX 1080Ti GPUs. 
After depth estimation, we reconstruct point clouds similar to MVSNet~\cite{yao_2018_mvsnet}. %

\subsection{Benchmark Performance}
\begin{table}
\setlength{\belowcaptionskip}{-0.5cm}
\setlength{\abovecaptionskip}{0.1cm}
\centering
\footnotesize
\begin{widetable}{\columnwidth}{cccc}
 \hline
 Methods & Acc.(mm) & Comp.(mm) & Overall(mm)\\
 \hline
 Camp~\cite{camp} & 0.835 & 0.554 & 0.695 \\ 
 Furu~\cite{furukawa2010} & 0.613 & 0.941 & 0.777 \\
 Tola~\cite{tola2012efficient} & 0.342 & 1.190 & 0.766 \\
 Gipuma~\cite{galliani_2015_gipuma} & \textbf{0.283} & 0.873 & 0.578 \\
 SurfaceNet~\cite{ji_2017_surfacenet}  & 0.450 & 1.040 & 0.745 \\
 MVSNet~\cite{yao_2018_mvsnet} & 0.396 & 0.527 & 0.462\\
 R-MVSNet~\cite{yao_2019_rmvsnet} & 0.383 & 0.452 & 0.417\\
 CIDER~\cite{xu2020learning_inverse} & 0.417 & 0.437 & 0.427\\
 P-MVSNet~\cite{luo_2019_p_mvsnet} & 0.406 & 0.434 & 0.420\\
 Point-MVSNet~\cite{chen_2019_pointmvsnet} & 0.342 & 0.411 & 0.376\\
 Fast-MVSNet~\cite{yu_2020_fastmvsnet} & 0.336 & 0.403 & 0.370\\ 
 CasMVSNet~\cite{gu_2020_cascademvsnet} & 0.325 & 0.385 & 0.355\\
 UCS-Net~\cite{cheng_2020_ucsnet} & 0.338 & 0.349 & \textbf{0.344} \\
 CVP-MVSNet~\cite{yang_2020_cvpr} & 0.296 & 0.406 & 0.351 \\
 Ours & 0.427 & \textbf{0.277} & 0.352\\
 \hline
\end{widetable}
\caption{Quantitative results of different methods on DTU's evaluation set~\cite{aanaes2016_dtu} (lower is better). }
\label{tab:evaluation_dtu}
\end{table}
\customparagraph{Evaluation on DTU Dataset}
We input images at their original size ($1600 \!\times\! 1200$) and set the number of views $N$ to 5.
The depth range for sampling depth hypotheses is fixed to $[425\,mm, 935\,mm]$. %
We follow the evaluation metrics provided by the DTU dataset~\cite{aanaes2016_dtu}. %
As shown in Table~\ref{tab:evaluation_dtu}, while Gipuma~\cite{galliani_2015_gipuma} performs best in \textit{accuracy}, 
our method outperforms others in \textit{completeness} and achieves competitive performance in \textit{overall quality}. 
Fig.~\ref{fig:dtu_qualitative} shows qualitative results. %
Our solution reconstructs a denser point cloud with finer details, which reflects in a high completeness. %
Further, our reconstruction at boundaries and thin structures appears better than of CasMVSNet~\cite{gu_2020_cascademvsnet}. 
Our adaptive propagation can recover from errors at boundaries, induced at coarser resolutions, by using the information of neighbors inside the boundary (\cf Fig.~\ref{fig:propagation}), while solely relying on sampling around a previous estimation, as CasMVSNet, can fail.

\begin{figure}
\centering
\setlength{\abovecaptionskip}{0.1cm}
\setlength{\belowcaptionskip}{-0.6cm}

{\includegraphics[width=1\linewidth]{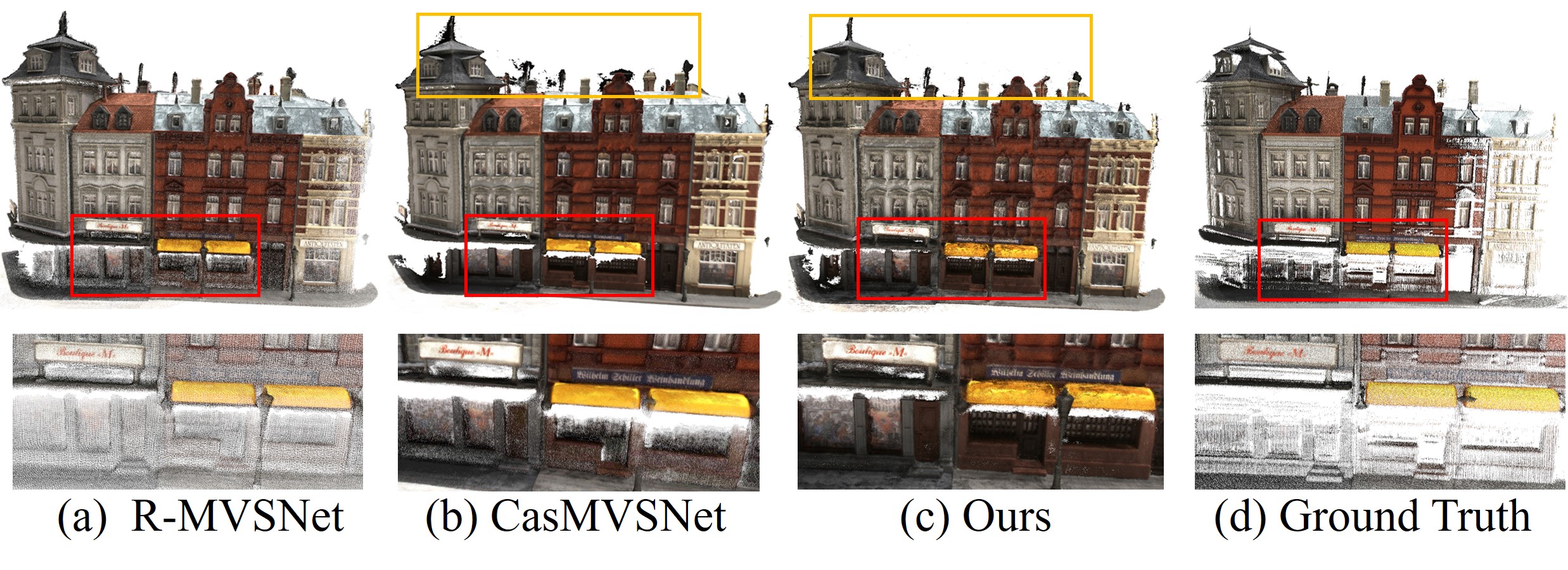}}
\caption{
    Qualitative comparison of scan 9 of DTU~\cite{aanaes2016_dtu}. 
    \textit{Top}: Reconstructed and ground truth point clouds. 
	Our method preserves the thin structures on the roof better than CasMVSNet~\cite{gu_2020_cascademvsnet} and delivers accurate boundaries.
    \textit{Bottom}: Zoom in. Capable of handling a high input resolution, our result is much denser, 
    with finer details at doors, windows and logos. 
    }
\label{fig:dtu_qualitative}
\end{figure}

\begin{figure}
\centering
\setlength{\belowcaptionskip}{-0.035cm}
\setlength{\abovecaptionskip}{0.1cm}
{\includegraphics[width=1.0\linewidth]{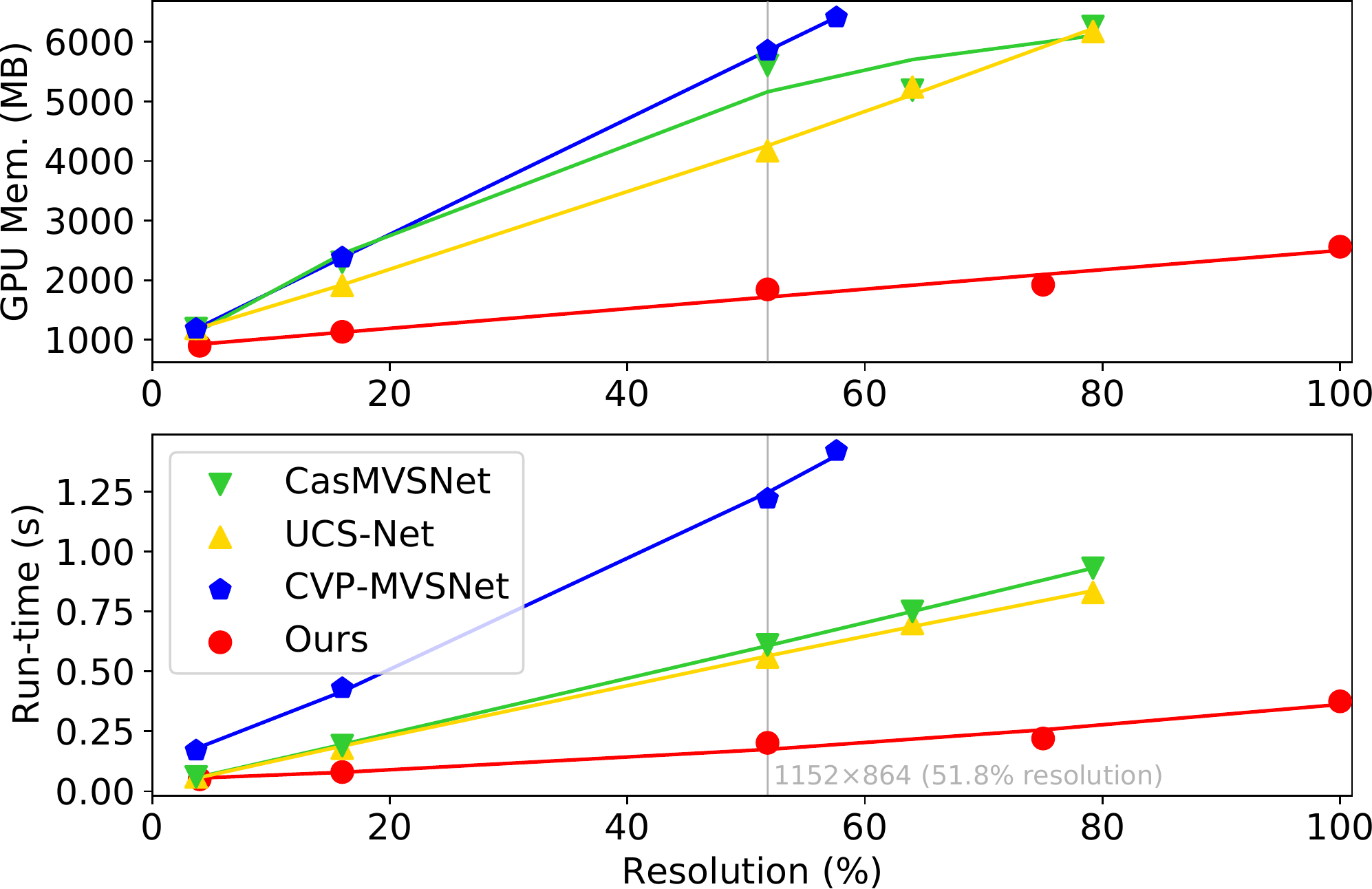}}
\caption{
    Relating GPU memory and run-time  to the input resolution on DTU's evaluation set~\cite{aanaes2016_dtu}. 
    The original image resolution is $1600\!\times\!1200$ (100\%). 
    Note that at higher resolutions other methods could not fit into the memory of a Nvidia RTX 2080 GPU, which is used for evaluation.
    }
\label{fig:memory_time_compare}
\end{figure}

\customparagraph{Memory and Run-time Comparison}
We compare the memory consumption and run-time with several state-of-the-art learning-based methods that achieve competing performance with low memory consumption and run-time: 
CasMVSNet~\cite{gu_2020_cascademvsnet}, UCS-Net~\cite{cheng_2020_ucsnet} and CVP-MVSNet~\cite{yang_2020_cvpr}. 
These methods propose a cascade formulation of 3D cost volumes and output depth maps at the same resolution as the input images. 
As shown in Fig.~\ref{fig:memory_time_compare}, memory and run-time of all the methods increase almost linearly w.r.t. the resolution as the number of depth hypotheses is fixed (notably this will lead to under-sampling of the enlarged image space for methods using a naive cost volume approach). Note that at higher resolutions other methods could not fit into the memory of the GPU used for evaluation.
We further observe that memory consumption and run-time increase much slower for PatchmatchNet than for other methods. 
For example, at a resolution of $1152 \!\times\! 864$ (51.8\%), 
memory consumption and run-time are reduced by 67.1\% and 66.9\% compared to CasMVSNet, 
by 55.8\% and 63.9\% compared to UCS-Net and 
by 68.5\% and 83.4\% compared to CVP-MVSNet. 
Combining the results in Table~\ref{tab:evaluation_dtu}, we conclude that our method is much more efficient 
in memory consumption and run-time than most state-of-the-art learning-based methods, at a very competitive performance.

\customparagraph{Evaluation on Tanks \& Temples Dataset}
We use the model trained on DTU~\cite{aanaes2016_dtu} without any fine-tuning. For evaluation, we set the input image size to $1920 \times 1056$ and the number of views $N$ to 7.
The camera parameters and sparse point cloud are recovered with OpenMVG~\cite{moulon2016openmvg}. %
During evaluation, the GPU memory and run-time for each depth map are 2887~MB and 0.505~s respectively. 
As shown in Table~\ref{tab:evaluation_tank}, the performance of our method on the intermediate dataset is comparable to CasMVSNet~\cite{gu_2020_cascademvsnet}, which has the highest score.
For the more complex advanced dataset, our method performs best among all the methods. 
Overall, due to its simple, scalable structure, our PatchmatchNet demonstrates competitive generalization performance, 
low memory consumption and low run-time compared to state-of-the-art learning-based methods that commonly use 3D cost volume regularization. 

\begin{table}
\setlength{\belowcaptionskip}{-0.5cm}
\setlength{\abovecaptionskip}{0.1cm}
\centering
\footnotesize
 \begin{widetable}{\columnwidth}{ccc}
 \hline
 Methods & Intermediate & Advanced\\
 \hline
 COLMAP~\cite{schoenberger2016colmap} & 42.14 & 27.24 \\ 
 MVSNet~\cite{yao_2018_mvsnet} & 43.48 & -\\
 R-MVSNet~\cite{yao_2019_rmvsnet} & 48.40 & 24.91\\
 CIDER~\cite{xu2020learning_inverse} & 46.76 & 23.12\\
 P-MVSNet~\cite{luo_2019_p_mvsnet} & 55.62 & -\\
 Point-MVSNet~\cite{chen_2019_pointmvsnet} & 48.27 & -\\
 Fast-MVSNet~\cite{yu_2020_fastmvsnet} & 47.39 & - \\
 CasMVSNet~\cite{gu_2020_cascademvsnet} & \textbf{56.42} & 31.12\\
 UCS-Net~\cite{cheng_2020_ucsnet} & 54.83 & - \\
 CVP-MVSNet~\cite{yang_2020_cvpr} & 54.03 & - \\
 Ours & 53.15 & \textbf{32.31}\\
 \hline
\end{widetable}

\caption{Results of different methods on Tanks \& Temples~\cite{knapitsch2017tanks}  ($F$ score, higher is better). 
Note that most methods refrain to evaluate on the more challenging \textit{Advanced} dataset. 
}
\label{tab:evaluation_tank}
\end{table}

\customparagraph{Evaluation on ETH3D Benchmark}
We use the model trained on DTU~\cite{aanaes2016_dtu} without any fine-tuning.
For evaluation, we set the input image size as $2688 \!\times\! 1792$. 
Due to the strong viewpoint changes in ETH3D, we also use $N\!=\!7$ views to utilize more multi-view information. 
Camera parameters and the sparse point cloud are recovered with COLMAP~\cite{schoenberger2016colmap}. %
During evaluation, the GPU memory consumption and run-time for the estimation of each depth map are 5529~MB and 1.250~s respectively. 
As shown in Table~\ref{tab:evaluation_eth}, on the training dataset, 
the performance of our method is comparable to COLMAP~\cite{schoenberger2016colmap} and PVSNet~\cite{xu2020pvsnetpv}. 
On the particularly challenging test dataset, our method performs best among all methods. 
Furthermore, our method is the fastest one evaluated so far (November 16th, 2020) on the ETH3D benchmark. 
Noting that PVSNet is a state-of-the-art learning-based method, 
the quantitative results demonstrate the effectiveness, efficiency and generalization capabilities of our method.

\begin{figure} [tbp]
\vspace{-0.2cm}
\setlength{\belowcaptionskip}{-0.2cm}
\centering
\setlength{\abovecaptionskip}{0.cm}
{\includegraphics[width=1.0\linewidth]{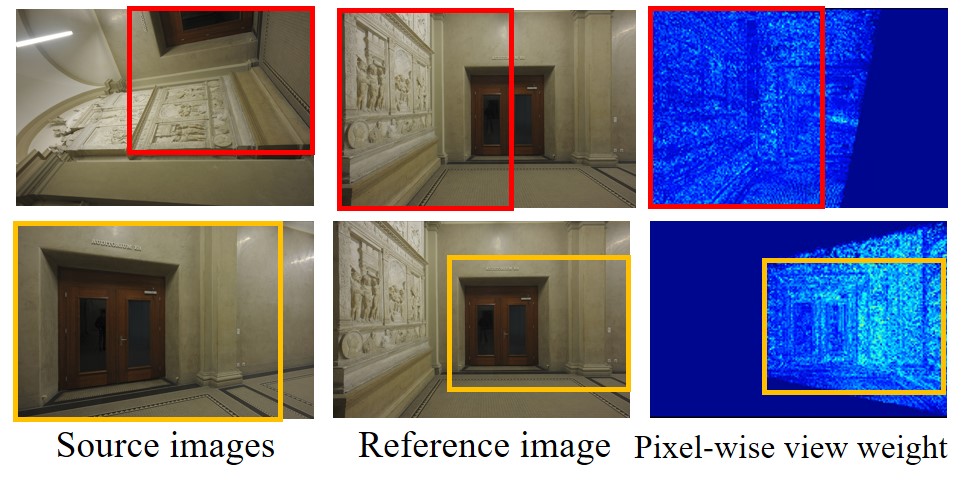}}
\caption{Visualization of our pixel-wise view weight on a scene from ETH3D~\cite{2017eth3d}. Areas marked with boxes in source images and reference image are co-visible. 
\emph{Right}: The corresponding pixel-wise view weights, 
bright colors (large values) indicate co-visibility. 
}
\label{fig:pixel_wise_view_weight}
\end{figure}

We visualize the pixel-wise view weight in  Fig.~\ref{fig:pixel_wise_view_weight}. 
Brighter colors indicate co-visible areas, \ie regions in the reference image that are also (well) visible in the source images. 
Conversely, areas that are not visible in the source images receive a dark color, corresponding to a low weight. 
Also pixel near depth discontinuities appear slightly darker than surrounding areas. 
By inspection, we conclude that our pixel-wise view weight is indeed 
capable to determine co-visible areas between the reference and source views. 

\begin{table}
\footnotesize
\centering
\setlength{\belowcaptionskip}{-0.5cm}
\setlength{\abovecaptionskip}{0.1cm}
\begin{widetable}{\columnwidth}{c|c|c|c|c}
 \hline
 \multirow{2}{*}{Methods} & \multicolumn{2}{c|}{Training}  & \multicolumn{2}{c}{Test}\\
 \cline{2-5}
  & $F_1$ score & Time(s) & $F_1$ score & Time(s)\\
  \hline
 MVE~\cite{fuhrmann2014mve} & 20.47 & 13278.69 & 30.37 & 10550.67\\
 Gipuma~\cite{galliani_2015_gipuma} & 36.38 & 587.77 & 45.18 & 689.75\\ 
 PMVS~\cite{furukawa2010} & 46.06 & 836.66 & 44.16 & 957.08\\
 COLMAP~\cite{schoenberger2016colmap} & \textbf{67.66} & 2690.62 & 73.01 & 1658.33 \\ 
 PVSNet~\cite{xu2020pvsnetpv} & 67.48 & - & 72.08 & 829.56 \\ 
 Ours & 64.21 & \textbf{452.63} & \textbf{73.12} & \textbf{492.52} \\
 \hline
\end{widetable}
\caption{
    Results of different methods on ETH3D~\cite{2017eth3d} ($F_1$ score, higher is better). 
    Due to strong viewpoint variations, currently, the only competitive pure learning-based method submitted on ETH3D is PVSNet~\cite{xu2020pvsnetpv}. 
    }
\label{tab:evaluation_eth}
\end{table}

\subsection{Ablation Study}
We conduct an ablation study to analyze the components. 
Unless specified, all the following studies are done on DTU's evaluation set~\cite{aanaes2016_dtu}. 

\customparagraph{Adaptive Propagation (AP) \& Adaptive Evaluation (AE)}
We compare our base model with versions that employ 
fixed 2D offsets, similar to Gipuma~\cite{galliani_2015_gipuma}, for propagation (w/o AP), and, 
fixed 2D offsets to sample the neighbors for spatial cost aggregation in the evaluation step (w/o AE).
As shown in Table~\ref{tab:ablation_adaptive_propa_eval}, 
our adaptive propagation and adaptive evaluation modules each improve results \wrt accuracy as well as completeness.

\begin{table}
\setlength{\belowcaptionskip}{-0.2cm}
\setlength{\abovecaptionskip}{0.1cm}
\centering
 \footnotesize
 \begin{widetable}{\columnwidth}{cccc}
 \hline
 Methods & Acc.(mm) & Comp.(mm) & Overall(mm)\\
 \hline
 w/o AP \& AE & 0.453 & 0.339 & 0.396 \\ 
 w/o AP & 0.437 & 0.285 & 0.361 \\
 w/o AE & 0.437 & 0.324 & 0.380 \\
 Ours & \textbf{0.427} & \textbf{0.277} & \textbf{0.352}\\
 \hline
\end{widetable}
\caption{Parameter sensitivity on DTU~\cite{aanaes2016_dtu} for Adaptive Propagation~(AP) and Adaptive Evaluation~(AE).
}
\label{tab:ablation_adaptive_propa_eval}
\end{table}

\customparagraph{Number of Iterations of Patchmatch}
Recall that, during training (\Sec \ref{implementation}), we do not include adaptive propagation for Patchmatch on stage 1. 
Consequently, we also keep the number of iterations on stage 1 as 1 for this investigation. 
More iterations of Patchmatch generally improve the performance, yet, the improvements stagnate after `2,2,1' iterations, Table~\ref{tab:ablation_iteration_num}. 
We further visualize the distribution of normalized absolute error in the inverse depth 
range for the setting `2,2,1' in Fig.~\ref{fig:error_distribution}. 
We observe that the error converges after only 5 iterations of Patchmatch across all stages. 
Compared to Gipuma~\cite{galliani_2015_gipuma} that employs a large number of neighbors for propagation, 
we have embedded Patchmatch in a coarse-to-fine framework to speed up long range interactions. 
Apart from that, our learned adaptive propagation, diverse initialization and local perturbation all contribute towards a faster converge.

\begin{table}
\setlength{\belowcaptionskip}{-0.3cm}
\setlength{\abovecaptionskip}{0.1cm}
\centering
\footnotesize
 \begin{widetable}{\columnwidth}{cccc}
 \hline
 Iterations & Acc.(mm) & Comp.(mm) & Overall(mm)\\

 \hline
 1,1,1 & 0.446 & 0.278 & 0.362 \\ 
 2,2,1 & 0.427 & \textbf{0.277} & 0.352 \\
 3,3,1 & \textbf{0.425} & 0.277 & \textbf{0.351} \\
 4,4,1 & 0.425 & 0.277 & 0.351 \\
 5,5,1 & 0.425 & 0.277 & 0.351 \\
 \hline
\end{widetable}

\caption{
    Ablation study of the number of Patchmatch iterations on DTU~\cite{aanaes2016_dtu}. %
    \textit{Iterations}, \textit{`a,b,c'} means that there are \textit{a}, \textit{b} and 
    \textit{c} iterations on stage 3, 2 and 1.
    Thanks to the coarse-to-fine framework, learned adaptive propagation, diverse initialization and local perturbation, our method converges after only 5 iterations combined on all stages.
    }
\label{tab:ablation_iteration_num}
\end{table}

\begin{figure}
\centering
\setlength{\belowcaptionskip}{-0.5cm}
\setlength{\abovecaptionskip}{0.1cm}
{\includegraphics[width=1.0\linewidth]{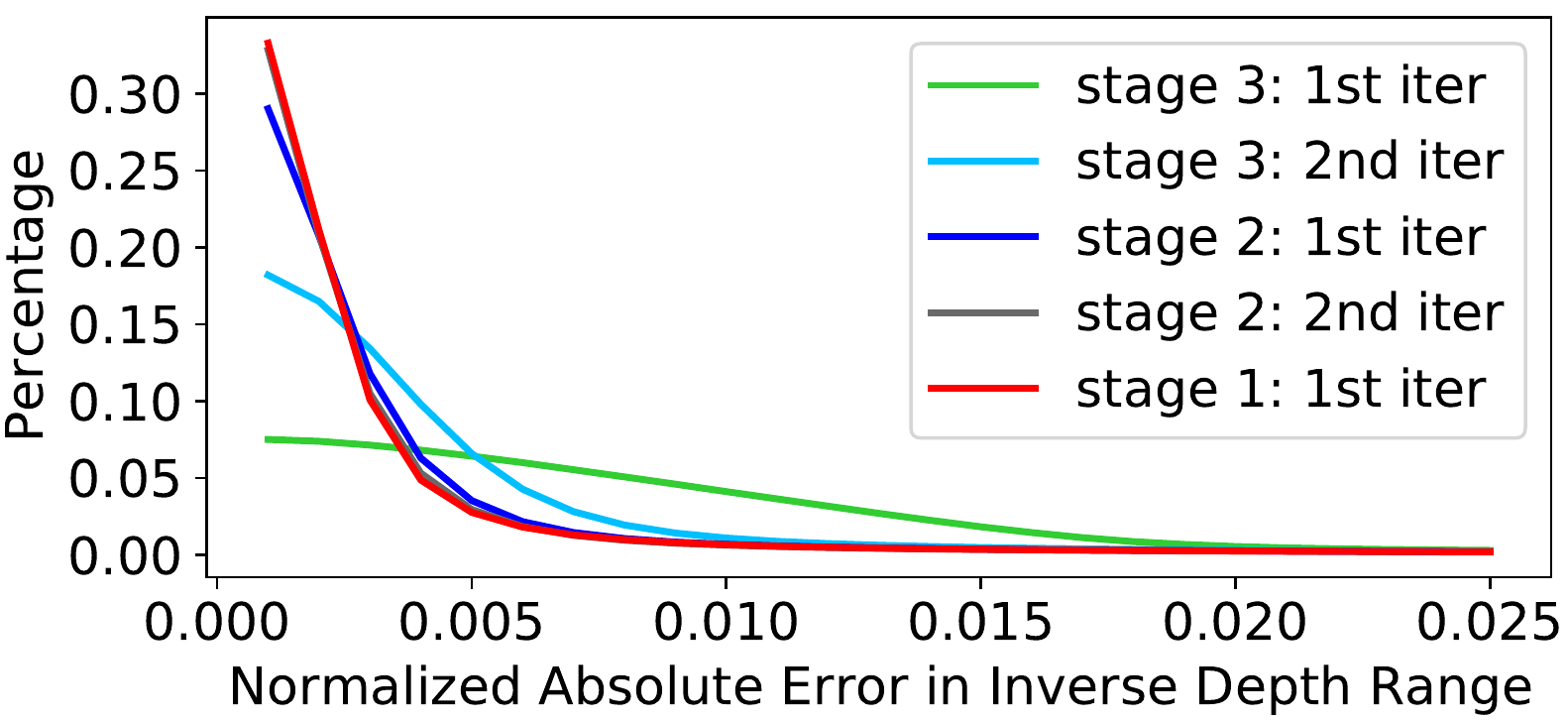}}
\caption{
    Distribution of normalized absolute errors in the inverse depth range on DTU's evaluation set~\cite{aanaes2016_dtu}.
    `\textit{Stage $k$, $n^{\textrm{th}}$ iter}' denotes the result of the $n^{\textrm{th}}$ iteration of Patchmatch on stage $k$. 
    Already after stage 3 our estimate is close to the solution while stage 2 and 1 refine it even more.
    }
\label{fig:error_distribution}
\end{figure}
\customparagraph{Pixel-wise View Weight~(VW) \& Robust Training Strategy~(RT)}
In this experiment, we resign from pixel-wise view weighting (w/o VW) and do not follow our strategy %
but choose four best source views for training (w/o RT). 
To investigate the generalization performance, we further test on Tanks \& Temples~\cite{knapitsch2017tanks} and ETH3D~\cite{2017eth3d}. 
Table~\ref{tab:ablation_view_weight_robust_training} shows a similar performance on DTU~\cite{aanaes2016_dtu} for all the models, yet, 
we observe a drop in performance on the other datasets without pixel-wise view weight or the robust training strategy. 
This proves that these two modules lead to improved robustness and a better generalization performance. 

\begin{table}
\setlength{\belowcaptionskip}{-0.15cm}
\setlength{\abovecaptionskip}{0.1cm}
\centering
 \footnotesize
 \begin{widetable}{\columnwidth}{cccc}
 \hline
 Methods & \makecell[c]{DTU} & \makecell[c]{Tanks\&Temples\\Intermediate} & \makecell[c]{ETH3D\\Training} \\
 \hline
 w/o VW \& RT & 0.351 & 52.05 & 60.40  \\ 
 w/o VW & 0.353 & 52.46 & 61.32  \\
 w/o RT & \textbf{0.348} & 52.12 & 62.57  \\
 Ours & 0.352 & \textbf{53.15} & \textbf{64.21} \\
 \hline
\end{widetable}
\caption{
    Ablation study concerning the pixel-wise view weight (VW) and the robust training strategy (RT).
    }
\label{tab:ablation_view_weight_robust_training}
\end{table}

\customparagraph{Number of Views}
Apart form our standard setting of $N=5$ views for DTU's evaluation set~\cite{aanaes2016_dtu},
we also evaluate the performance when $N=2,3,6$. 
Using more views is known to improve performance, \eg by alleviating the occlusion problem, 
which coincides with our findings summarized in Table~\ref{tab:ablation_number_of_views}. 
With more input views, the reconstruction quality tends to improve in both accuracy and completeness. 

\begin{table}
\centering
\setlength{\belowcaptionskip}{-0.5cm}
\setlength{\abovecaptionskip}{0.1cm}
\footnotesize
 \begin{widetable}{\columnwidth}{cccc}
 \hline
 $N$ & Acc.(mm) & Comp.(mm) & Overall(mm)\\
 \hline
 2 & 0.439 & 0.332 & 0.385 \\ 
 3 & 0.428 & 0.284 & 0.356 \\
 5 & \textbf{0.427} & \textbf{0.277} & \textbf{0.352}\\
 6 & 0.429 & 0.278 & 0.353 \\
 \hline
\end{widetable}
\caption{
    Ablation study of the number of input views $N$ on DTU's evaluation set~\cite{aanaes2016_dtu}.
    }\label{tab:ablation_number_of_views}
\end{table}

%% file: tex/conclusion.tex
\section{Conclusion}
We present PatchmatchNet, a novel cascade formulation of learning-based Patchmatch, 
augmented with learned adaptive propagation and evaluation modules based on deep features.
Inherited from its name-giving ancestor, PatchmatchNet naturally possesses low memory requirements, 
independent of the disparity range and unlike most learning-based methods, PatchmatchNet does not rely on 3D cost volume regularization. 
Embedded into a cascade formulation, PatchmatchNet further shows a high processing speed. 
Despite its simple structure, extensive experiments on DTU, Tanks \& Temples and ETH3D demonstrate 
a remarkably low computation time, low memory consumption, favorable generalization properties and 
competitive performance compared to the state-of-the-art. %
PatchmatchNet makes learning-based MVS more efficient and more applicable to memory restricted devices or time critical applications. 
For the future, we hope to apply it on movable platforms such as mobile phones and head mounted displays, where the computation resource is limited.

%% file: tex/appendix.tex
\newpage
\begin{center}
      {\Large \bf Supplementary Material}
\end{center}
\setcounter{section}{0}

\section{Why not use 3D cost volume regularization?}\label{no_3d_cnn}
The adaptive evaluation of our learning-based Patchmatch utilizes 3D convolution layers with $1 \times\!1\!\times\!1$ 
kernels for the matching cost computation as well as the pixel-wise view weight estimation. 
This is in contrast to common previous works~\cite{yao_2018_mvsnet, luo_2019_p_mvsnet, xu2020learning_inverse, gu_2020_cascademvsnet, cheng_2020_ucsnet, yang_2020_cvpr, xu2020pvsnetpv} 
where a 3D U-Net regularizes the cost volume. 
Similarly, arguing that the distribution of cost volume itself being not discriminative enough~\cite{seki2016patch,fu2018learning}, 
PVSNet~\cite{xu2020pvsnetpv} also applies a 3D U-Net for predicting the visibility per source view. 

The problem with such regularization framework is that it requires a regular spatial structure in the volume. 
Although we concatenate the matching costs per pixel and depth hypothesis into a volume-like shape as other 
works~\cite{yao_2018_mvsnet, luo_2019_p_mvsnet,xu2020learning_inverse, gu_2020_cascademvsnet, cheng_2020_ucsnet, yang_2020_cvpr, xu2020pvsnetpv}, 
we do not possess such a regular structure: 
(\romannumeral 1) the depth hypotheses for each pixel and its spatial neighbors are different, 
which makes it difficult to aggregate cost information in the spatial domain; 
(\romannumeral 2) the depth hypotheses of each pixel are not uniformly distributed in the inverse depth range as 
CIDER~\cite{xu2020learning_inverse}, which makes it difficult to aggregate cost information along depth dimension. 

Recall that during the computation of the pixel-wise view weights in the initial iteration of Patchmatch, 
depth hypotheses are randomly distributed in the \textit{inverse} depth range, 
\ie the hypotheses are spatially different per pixel. %
In each subsequent iteration (on stage $k$), we perform local perturbation by generating per pixel $N_k$ depth hypotheses uniformly 
in the normalized inverse depth range $R_k$, which is centered at the previous estimate. 
Consequently, the hypotheses of spatial neighbors can differ significantly, especially at depth discontinuities and thin structures.
Including the hypotheses obtained by adaptive propagation, that are, moreover, not uniform in the inverse depth range, will increase these effects further. 

In the end, however, the main reason for our approach to avoid 3D cost volume regularization altogether is efficiency. 
In a coarse-to-fine framework, running such regularization frameworks over multiple iterations of Patchmatch on each stage would increase memory consumption and run-time significantly and mitigate our main contribution of building a high-performance, but particularly lightweight framework that can operate with a high computation speed.

\section{How to set the normalized inverse depth range $R_k$ in the local perturbation step of Patchmatch?}
After the initial iteration, our set of hypothesis is obtained by adaptive propagation 
and by local perturbation of the previous estimation. 
Recall that our local perturbation procedure enriches the set of hypothesis 
by generating per pixel $N_k$ depth 
hypotheses uniformly in the normalized inverse depth range $R_k$. 
The objective is two-fold. 
Especially at the beginning, at low resolution, this helps to further explore the search space. 
More importantly, our adaptive propagation implicitly assumes front-to-parallel surfaces, 
since we do not explicitly include tangential surface information (due to an implied heavy memory consumption)
like~\cite{bleyer2011patchmatch, galliani_2015_gipuma, xu_2019_acmm}. %
Sampling in the local vicinity of the previous estimation will refine the solution locally and %
mitigate potential disadvantages from not explicitly modeling tangential surface information. 
We find it helpful to apply these perturbations already at an early stage to inject the positive effects into hypothesis propagation 
and note that a-posteriori refinement at the finest level alone cannot recover the same quality. %
In practice, we again operate in coarse-to-fine manner and set $R_k$ accordingly, based on the hierarchy level. 

\begin{figure}
\centering
\setlength{\belowcaptionskip}{-0.5cm}
\setlength{\abovecaptionskip}{0.1cm}
{\includegraphics[width=1.0\linewidth]{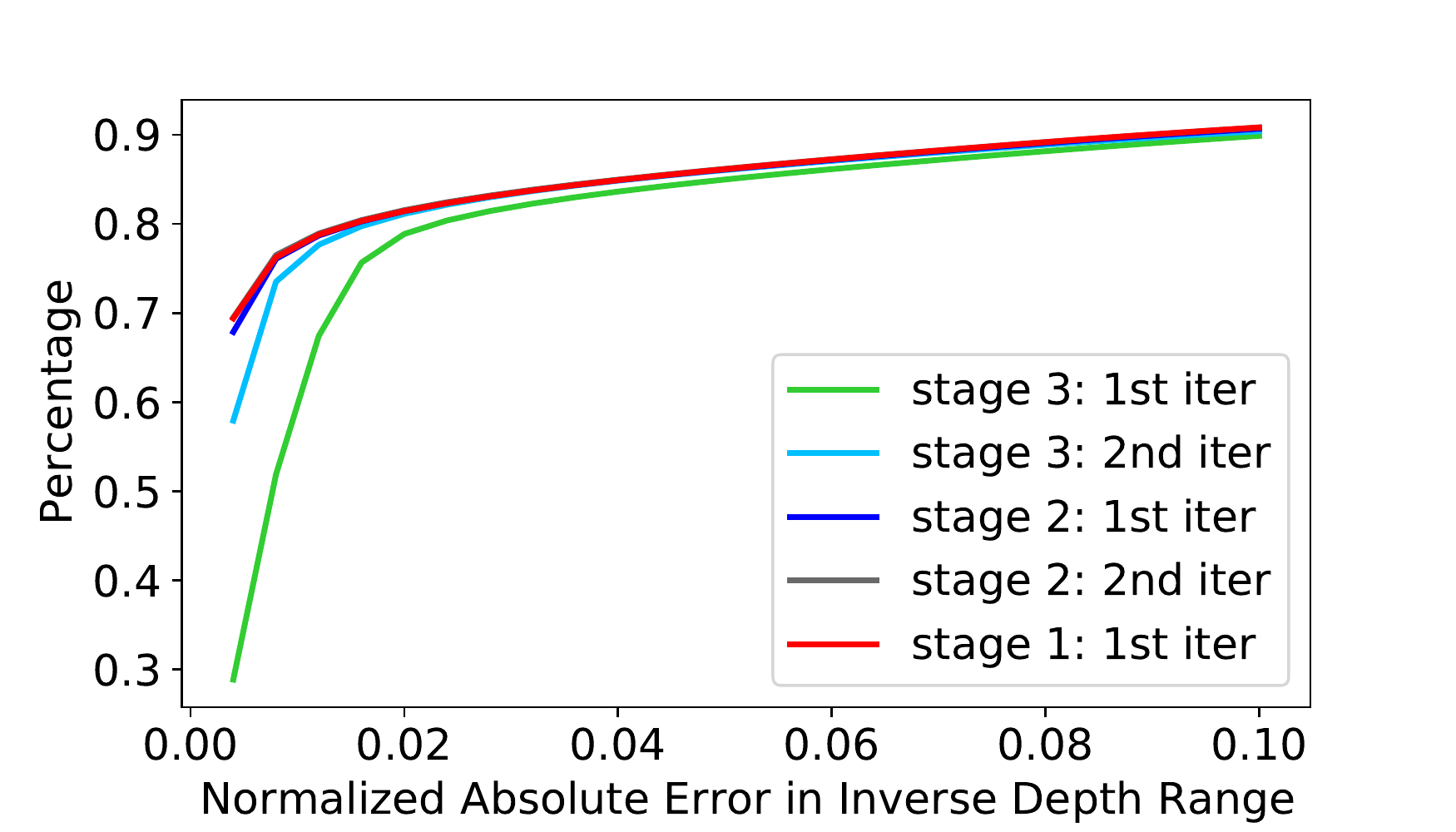}}
\caption{Cumulative distribution function of normalized absolute errors in the inverse depth range on DTU's evaluation set~\cite{aanaes2016_dtu}. `\textit{Stage $k$, $n^{\textrm{th}}$ iter}' denotes the result of the $n^{\textrm{th}}$ iteration of Patchmatch on stage $k$. 
}
\label{fig:depth_range_initialization}
\end{figure}

Fig.\ref{fig:depth_range_initialization} shows the cumulative distribution function of the normalized 
absolute error in the inverse depth range on DTU's evaluation set~\cite{aanaes2016_dtu}. 
After the first iteration of Patchmatch on stage 3, the estimation error decreases remarkably: 
the normalized error is already smaller than 0.1 for 90.0\% percent of the cases. %
Visibly, the performance keeps improving after each iteration. 
To correct errors in estimation and refine the results on stage $k$, we set $R_k$ to compensate most of estimation errors. For example, we set $R_3=0.38$ for Patchmatch on stage 3 after first iteration so that we can cover most ground truth depth in the hypothesis range and then refine the results. %
Besides, adaptive propagation will further correct those wrong estimations with the depth hypotheses from neighbors when sampling in $R_k$ fails in refinement (\cf Fig.~6 from the paper).

\section{Why not include propagation for last iteration of Patchmatch on stage 1?}
Similar to MVSNet~\cite{yao_2018_mvsnet}, the point cloud reconstruction mainly consists of photometric consistency filtering, 
geometric consistency filtering and depth fusion. %
Photometric consistency filtering is used to filter out those depth hypotheses that have low confidence. 
Based on MVSNet~\cite{yao_2018_mvsnet}, we define the confidence as the probability sum of the depth hypotheses that fall in a small range near the estimation. 
We use the probability $\mathbf{P}$ (\cf Eq.~7 from the paper) from the last iteration of Patchmatch on stage 1 for filtering. 
In this iteration, we only perform local perturbation, without adaptive propagation. 
At stage 1, operating at a quarter the image resolution and  %
with the algorithm almost converged, 
the hypotheses obtained via propagation from spatial neighbors are usually very similar to the current solution. %
Such irregular sampling of the probability space causes bias in the regression (\cf Eq.~7 from the paper) 
and the estimate becomes over-confident at the current solution, where most propagated samples are located. 
In contrast, by performing only the local perturbation, the depth hypotheses are uniformly distributed in the inverse depth range. 
Contrary to previous iterations, 
we compute the estimated depth at pixel $\mathbf{p}$, $\mathbf{D}(\mathbf{p})$,  by utilizing the inverse depth regression~\cite{xu2020learning_inverse}, which is based on the \textit{soft argmin} operation~\cite{kendall_2017_gcnet}:
\begin{equation}\label{eq:disp_regression}
\setlength{\abovedisplayskip}{3pt}
\setlength{\belowdisplayskip}{3pt}
    \mathbf{D}(\mathbf{p}) = (\sum_{j=0}^{D-1} \frac{1}{d_{j}} \cdot \mathbf{P}(\mathbf{p},j))^{-1},
\end{equation}
where $\mathbf{P}(\mathbf{p},j)$ is the probability for pixel $\mathbf{p}$ at the $j$-th depth hypothesis. 
Then we compute the probability sum of four depth hypotheses that are nearest to the estimation to measure the confidence~\cite{yao_2018_mvsnet}.

\section{Weighting in the Adaptive Spatial Cost Aggregation}
Recall that in Eq.~6 of the paper we utilize two weights to aggregate our spatial costs, 
$\left\{ w_k \right\}_{k=1}^{K_{e}}$ based on spatial feature similarity and 
$\left\{ d_k \right\}_{k=1}^{K_{e}}$ based on the similarity of depth hypotheses. 
The feature weights $\left\{ w_k \right\}_{k=1}^{K_{e}}$ at a pixel $\mathbf{p}$ are based on the 
feature similarity at the sampling locations around $\mathbf{p}$, 
measured in the reference feature map $\mathbf{F}_0$. 
Given the sampling positions $\left\{ \mathbf{p}+\mathbf{p}_k+\Delta \mathbf{p}_k \right\}_{k=1}^{K_{e}}$, 
we extract the corresponding features from $\mathbf{F}_0$ via bilinear interpolation. 
Then we apply group-wise correlation~\cite{guo2019group} between the features at each sampling location and $\mathbf{p}$. 
The results are concatenated into a volume on which we apply 3D convolution layers with $1 \!\times\! 1 \!\times\! 1$ 
kernels and sigmoid non-linearities to output normalized weights that describe the similarity between each sampling point and $\mathbf{p}$.

As discussed in \Sec~\ref{no_3d_cnn}, neighboring pixels will be assigned different depth values throughout the estimation process. 
For pixel $\mathbf{p}$ and the $j$-th depth hypothesis, our depth weights $\left\{ d_k \right\}_{k=1}^{K_{e}}$ take this into account and downweight the influence 
of samples with large depth difference, especially when located across depth discontinuities. 
To that end, we collect the absolute difference in inverse depth between each sampling point and pixel $\mathbf{p}$ with their $j$-th hypotheses, 
and obtain the weights by applying a sigmoid function on the, again, inverted differences for normalization.

\section{Evaluation of Multi-stage Depth Estimation}
We use multiple stages to estimate the depth map in a coarse-to-fine manner. Here, we analyze the effectiveness of our multi-stage framework. We upsample the estimated depth maps on stages 3, 2 and 1, to the same resolution as the input and then reconstruct the point clouds. As shown in Table~\ref{tab:evaluation_multi_stage}, the reconstruction quality gradually increases from coarser stages to finer ones. %
This shows that our multi-stage framework can reconstruct the scene geometry with increasing accuracy and completeness. 

\begin{table}
\footnotesize
\centering
\setlength{\belowcaptionskip}{-0.2cm}
\setlength{\abovecaptionskip}{0.1cm}
\begin{widetable}{\columnwidth}{cccc}
 \hline
 Stages & Acc.(mm) & Comp.(mm) & Overall(mm)\\
 \hline
 3 & 0.740 & 0.389 & 0.564 \\ 
 2 & 0.471 & 0.283 & 0.377 \\
 1 & 0.441 & \textbf{0.268} & 0.354 \\
 0 & \textbf{0.427} & 0.277 & \textbf{0.352}\\
 \hline
\end{widetable}

\caption{Quantitative results of different stages on DTU's evaluation set~\cite{aanaes2016_dtu} (lower is better). The depth maps on stages 3, 2 and 1 are upsampled to reach the same resolution as input images and then used to reconstruct point clouds.}
\label{tab:evaluation_multi_stage}
\end{table}

\begin{figure}[t]
\centering
\setlength{\belowcaptionskip}{-0.3cm}
\setlength{\abovecaptionskip}{0.1cm}
{\includegraphics[width=1.0\linewidth]{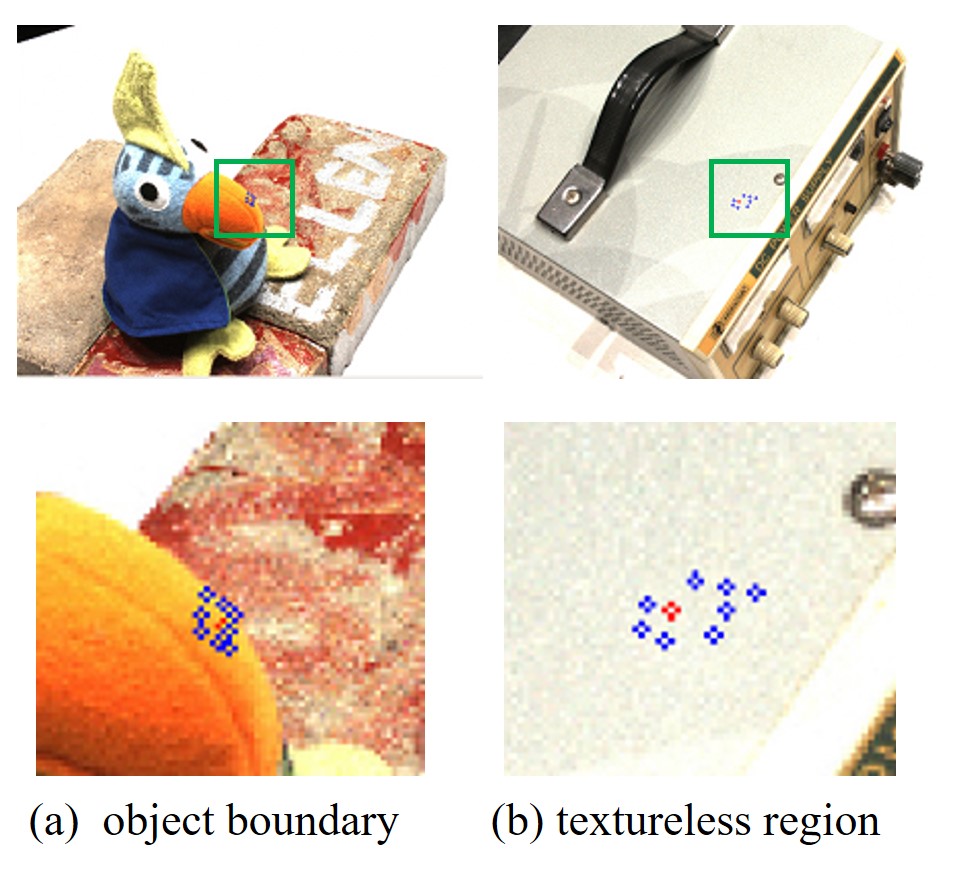}}
\caption{Visualization of sampling locations in adaptive propagation for two typical situations: object boundary and textureless region. The center points and sampling points are shown in red and blue respectively. %
}
\label{fig:propagation_result}
\end{figure}

\begin{figure}[t]
\centering
\setlength{\belowcaptionskip}{-0.5cm}
\setlength{\abovecaptionskip}{0.1cm}
{\includegraphics[width=1.0\linewidth]{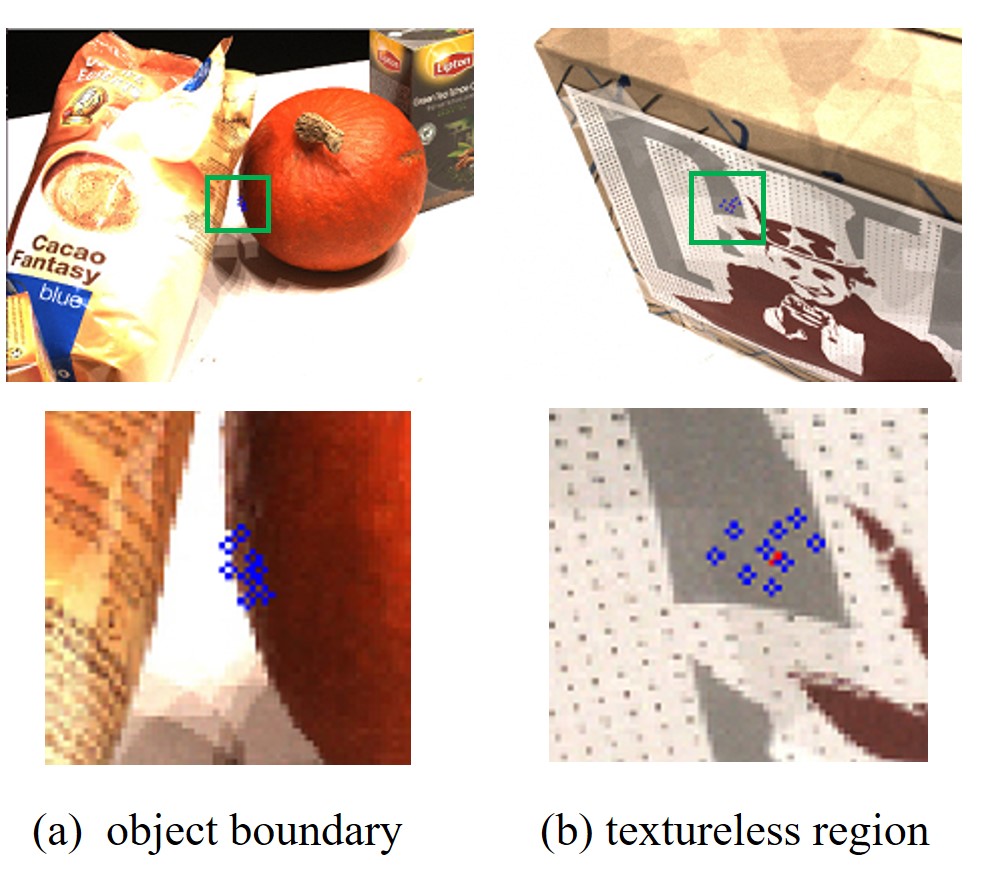}}
\caption{Visualization of sampling locations in adaptive evaluation for two typical situations: object boundary and textureless region. The center points and sampling points are shown in red and blue respectively. %
}
\label{fig:evaluation_result}
\end{figure}

\section{Visualization of Adaptive Propagation}
We visualize the sampling locations in two typical situations, at an object boundary and a textureless region. %
As shown in Fig.~\ref{fig:propagation_result}, for the pixel $\mathbf{p}$ at the object boundary, 
all sampling points tend to be located on the same surface as $\mathbf{p}$. %
In contrast, for the pixel $\mathbf{q}$ in the textureless region, the sampling points are spread over a larger region. 
By sampling from a large region, a more diverse set of depth hypotheses can be propagated to $\mathbf{q}$ 
and reduce the local ambiguity for depth estimation in the textureless area. 
The visualization shows two examples how the adaptive propagation successfully adapts the sampling to different challenging situations.

\section{Visualization of Adaptive Evaluation}
Here, we again visualize the sampling locations for two situations, at an object boundary and a textureless region. %
Fig.\ref{fig:evaluation_result} demonstrates that for the pixel $\mathbf{p}$ at the object boundary, 
sampling points tend to stay within the boundaries of the object, such that they focus on similar depth regions. %
For the pixel $\mathbf{q}$ in the textureless region, the points are distributed sparsely to sample from a large context, 
which helps to obtain reliable matching and to reduce the ambiguity. 
Again, the visualization demonstrates how our adaptive evaluation adapts the sampling for the spatial cost aggregation to different situations.

\if{0}
\section{Visualization of Pixel-wise View Weight}
Since ETH3D benchmark~\cite{2017eth3d} has strong viewpoint changes, we visualize the pixel-wise view weight to qualitatively analyze whether it captures visibility information. As shown in Fig.\ref{fig:pixel_wise_view_weight}, for areas in the reference image that are visible in the source image, the corresponding pixel-wise view weights are positive, which indicates that these are co-visible areas. In contrast, for the areas in the reference image that are invisible in the source image, the weights are almost zero, which indicates that they are invisible. So we can conclude that the pixel-wise view weight can reasonably determine the co-visible areas in the reference view and source views. By considering visibility information while aggregating matching costs for all source views, the influence of noise can be reduced and thus improve the accuracy. 

\begin{figure}
\centering
{\includegraphics[width=1\linewidth]{figures/pixel_wise_view_weight_2.png}}
\caption{Visualization of pixel-wise view weight on ETH3D benchmark~\cite{2017eth3d}. \textbf{Top row}: source images. \textbf{Middle Row}: reference image. \textbf{Bottom row}: pixel-wise view weight. For each source image, the area marked with red box in the reference image is visible in the source image (the corresponding area is marked with green box). In the corresponding pixel-wise view weight, this area has positive values, which indicates it is visible. 
}
\label{fig:pixel_wise_view_weight}
\end{figure}

\fi

\section{Visualization of Point Clouds}
\begin{figure*}
\centering
{\includegraphics[width=1.0\linewidth]{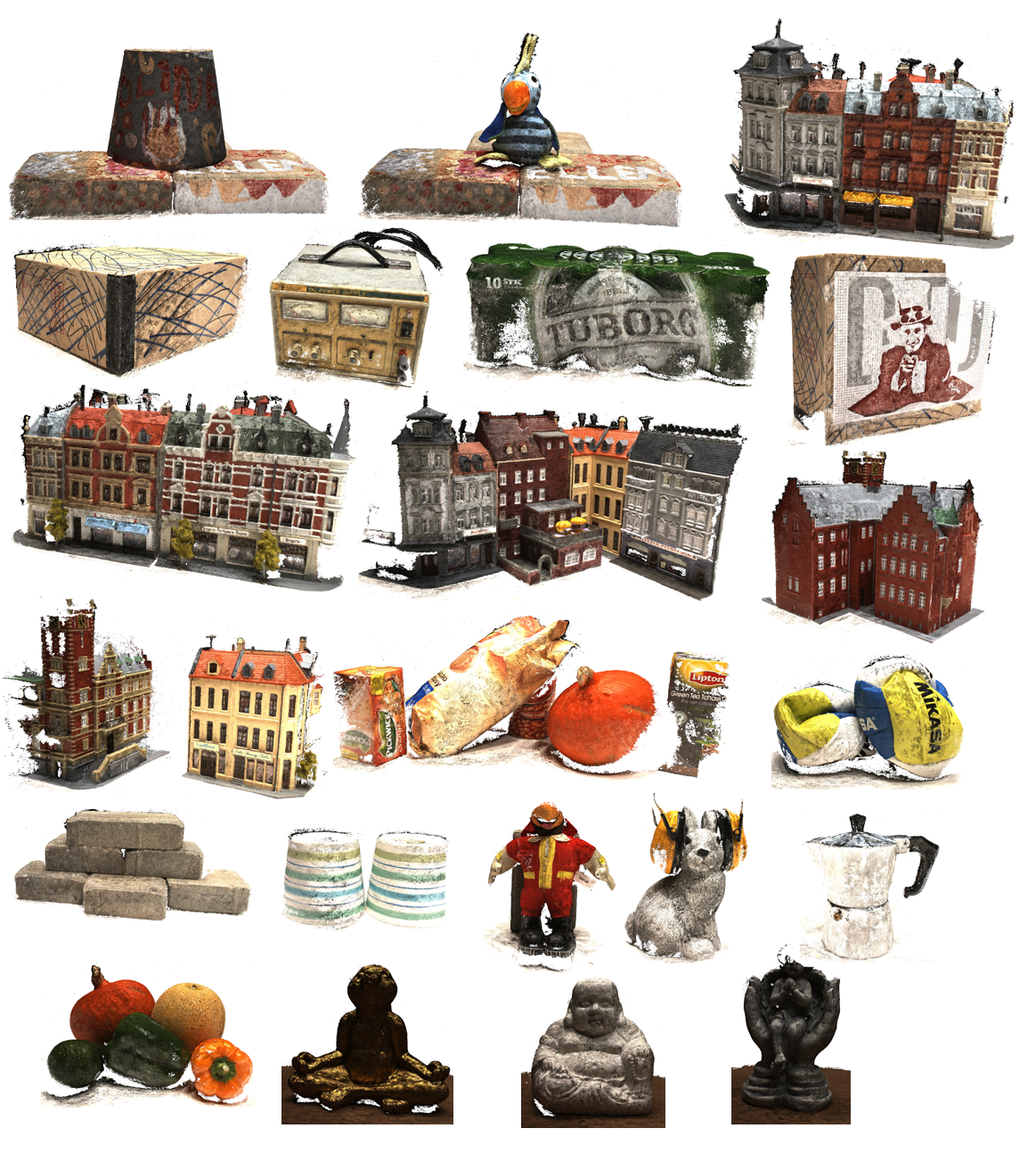}}
\caption{Reconstruction results on DTU's evaluation set~~\cite{aanaes2016_dtu}.}
\label{fig:dtu_qualitative_all}
\end{figure*}

\begin{figure*}
\centering
{\includegraphics[width=1.0\linewidth]{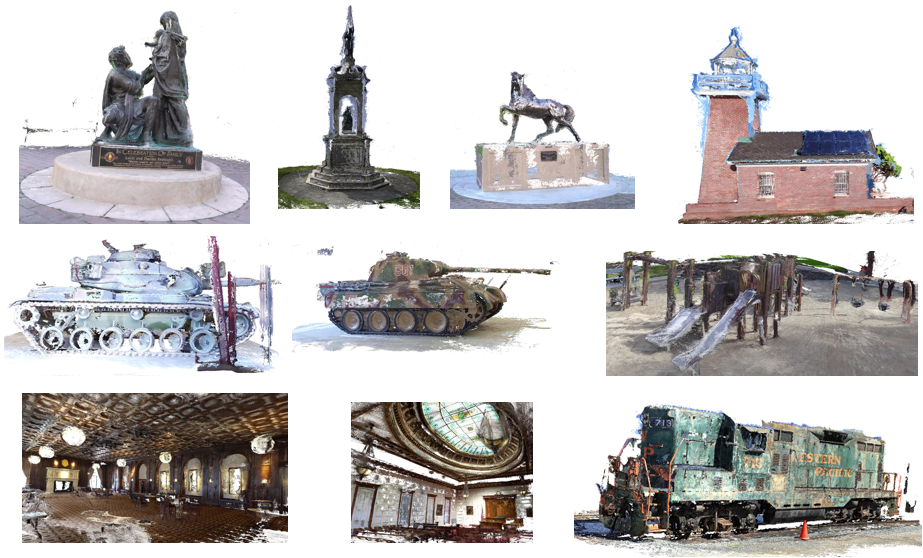}}
\caption{Reconstruction results on Tanks \& Temples dataset~\cite{knapitsch2017tanks}.}
\label{fig:tanks_qualitative}
\end{figure*}

\begin{figure*}
\centering
{\includegraphics[width=1.0\linewidth]{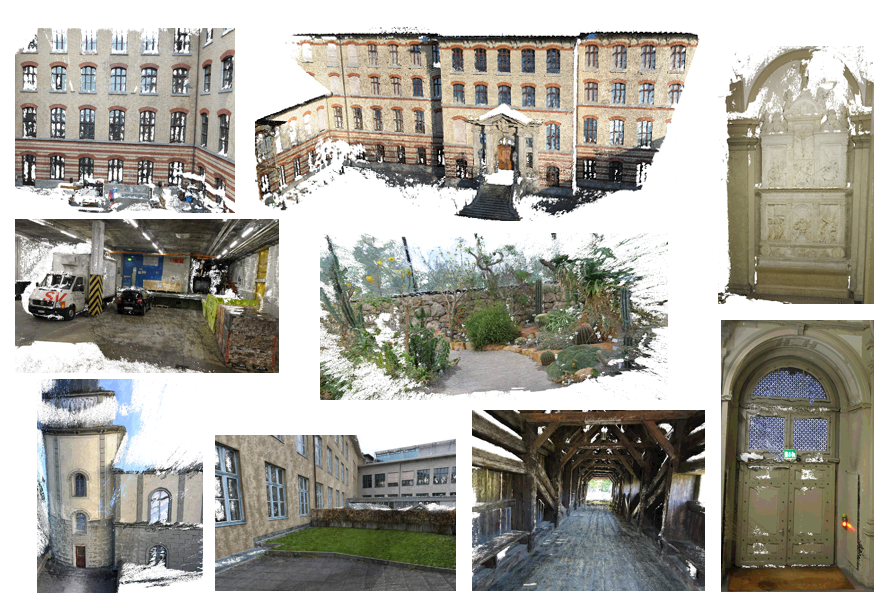}}
\caption{Reconstruction results on ETH3D benchmark~\cite{2017eth3d}.}
\label{fig:eth_qualitative}
\end{figure*}

We visualize reconstructed point clouds from DTU's evaluation set~\cite{aanaes2016_dtu}, Tanks \& Temples dataset~\cite{knapitsch2017tanks} and ETH3D benchmark~\cite{2017eth3d}  in Fig.~\ref{fig:dtu_qualitative_all}, ~\ref{fig:tanks_qualitative}, ~\ref{fig:eth_qualitative}. 
\clearpage

\clearpage

\clearpage

\clearpage